\documentclass{article} 

\usepackage{pdt_report}  
\usepackage{algorithm}
\usepackage{algorithmic}
\usepackage{amsmath}
\usepackage{amssymb}
\usepackage{amsthm}
\usepackage[toc,page]{appendix}
\usepackage{array}
\usepackage{bibunits}
\usepackage{booktabs} 
\usepackage{bm}
\usepackage{bbm}
\usepackage{cancel}
\usepackage{enumitem}
\usepackage{lipsum}
\usepackage{mathtools}
\usepackage[framemethod=default]{mdframed}
\usepackage{multicol}
\usepackage[most]{tcolorbox}
\usepackage{graphicx}
\usepackage{subcaption}
\usepackage{titletoc}
\usepackage{tikz}
\usepackage{wrapfig}
\usepackage[dvipsnames]{xcolor}
\usepackage[textsize=tiny]{todonotes}

\usetikzlibrary{bayesnet}
\usetikzlibrary{calc}

\definecolor{boilermakergold}{HTML}{cfb991}
\definecolor{steel}{HTML}{555960}
\definecolor{coolgray}{HTML}{6f727b}
\definecolor{dust}{HTML}{EBD99F}

\theoremstyle{plain}
\newtheorem{theorem}{Theorem}[section]
\newtheorem{proposition}[theorem]{Proposition}

\theoremstyle{definition}

\theoremstyle{remark}

\definecolor{softteal}{RGB}{70, 190, 180} 

\usepackage{pifont}

\def\eqref#1{equation~\ref{#1}}

\def\1{\bm{1}}

\def\rvu{{\mathbf{i}}}

\def\rvm{{\mathbf{m}}}

\def\rvo{{\mathbf{o}}}

\def\rvs{{\mathbf{s}}}

\def\rvu{{\mathbf{u}}}

\def\vtheta{{\bm{\theta}}}

\def\vb{{\bm{b}}}

\def\ve{{\bm{e}}}

\def\vm{{\bm{m}}}

\def\vo{{\bm{o}}}

\def\vr{{\bm{r}}}
\def\vs{{\bm{s}}}

\def\vx{{\bm{x}}}

\def\mI{{\bm{I}}}

\def\mSigma{{\bm{\Sigma}}}

\DeclareMathAlphabet{\mathsfit}{\encodingdefault}{\sfdefault}{m}{sl}
\SetMathAlphabet{\mathsfit}{bold}{\encodingdefault}{\sfdefault}{bx}{n}

\def\gD{{\mathcal{D}}}

\def\gH{{\mathcal{H}}}

\def\gK{{\mathcal{K}}}
\def\gL{{\mathcal{L}}}

\def\gN{{\mathcal{N}}}
\def\gO{{\mathcal{O}}}

\def\gQ{{\mathcal{Q}}}

\def\gS{{\mathcal{S}}}

\def\gU{{\mathcal{U}}}

\newcommand{\E}{\mathbb{E}}

\newcommand{\R}{\mathbb{R}}

\newcommand{\KL}{D_{\mathrm{KL}}}

\usepackage{makecell}

\DeclareMathOperator*{\argmax}{arg\,max}
\DeclareMathOperator*{\argmin}{arg\,min}

\providecommand{\divergencemean}{--}

\usepackage{cleveref}
\crefname{assumption}{assumption}{assumptions}
\Crefname{assumption}{Assumption}{Assumptions}
\crefname{corollary}{corollary}{corollaries}
\Crefname{corollary}{Corollary}{Corollaries}
\crefname{equation}{Eq.}{Eqs.}
\Crefname{equation}{Eq.}{Eqs.}

\title{%
  Controllable Sim Agents via Behavior Latents
}

\author{
  Juanwu~Lu \\
  Purdue University \\
  \texttt{juanwu@purdue.edu} \\
  \And
  Junyu~Zhu \\
  University of Tokyo \\
  \texttt{jimmyzhu@iis.u-tokyo.ac.jp} \\
  \And
  Ziran~Wang \\
  Purdue University \\
  \texttt{ziran@purdue.edu}
}

\definecolor{boilermakergold}{HTML}{cfb991}
\definecolor{steel}{HTML}{555960}
\definecolor{coolgray}{HTML}{6f727b}

\begin{document}

\maketitle

\begin{bibunit}[plainnat]
  \begin{abstract}
    \label{sec: abstract}
    Realistic traffic simulation requires agents that imitate logged behavior and can also be steered along interpretable axes. Such controllability enables engineers to isolate variables, reproduce specific edge cases, and test autonomous systems without real-world risk. We introduce Controllable Neural Variational Agents (CNeVA), a controllable simulated-agent framework that learns to infer a per-agent Gaussian behavior latent from per-channel discounted returns via a closed-form conjugate variational update, conditioning a rectified-flow trajectory generator trained on a mixed channel-mask curriculum for classifier-free guidance. To tackle scarcity in reward signals, we propose soft eligibility gates that replace hard binary thresholds with smooth exponential decay, preserving the gradient signal for near-threshold agents. On the Waymo Open Motion Dataset, CNeVA attains competitive realism on the benchmark while exposing per-channel controllability that the higher-ranked imitation models lack. Speed- and acceleration-based steering produces monotone responses without stall-induced reward hacking. Safety controllability is monotone and substantial with the introduction of soft eligibility. We manage to achieve steerable map compliance under a context-residual return measure. Furthermore, our experiment demonstrates that steering metrics must be read alongside physical-plausibility guardrails to avoid reward-hacking confounds.
\end{abstract}


  \section{Introduction}
\label{sec: introduction}

As autonomous vehicles (AVs) move toward large-scale deployment, ensuring their safety and reliability requires extensive validation. However, real-world testing is costly, while conventional traffic simulators remain limited in fidelity and scalability. Recent advances in generative models have made data-driven traffic simulation a promising alternative for AV testing, enabling simulated agents to both reproduce realistic behaviors and respond to AV actions. For adversarial evaluation, simulated agents are often expected to exhibit specific behavioral preferences as well. Although imitation learning can enable the generation of diverse and realistic traffic behaviors~\citep{seff2023motionlm,wu2024smart,philion2024trajeglish,zhang2025catk}, it generally lacks fine-grained controllability over agent behavior and therefore struggles to adapt generated scenarios to user-specified objectives.

Generative modeling has expanded the diversity of synthesized traffic scenarios. Existing methods using variational autoencoders (VAE)~\citep{suo2021trafficsim,zhang2023trafficbots,liu2026stage} produce behavioral variation through latent-space resampling but cannot deliberately steer behavior toward specific driving preferences. Diffusion-based models~\citep{jiang2023motiondiffuser,cao2024rlhf,jiang2024scenediffuser,tan2025scenediffuser2} improve controllability by injecting guidance during the generation process, yet achieving stylistic trajectories through guidance tuning is often nontrivial in practice, and excessive guidance can lead to unrealistic trajectories. Self-play reinforcement learning encourages desired behaviors through hand-crafted reward functions in closed-loop interaction~\citep{cornelisse2024hrppo,cusumano2025gigaflow,chang2026spacer}. Despite their controllability, these methods rely on computationally expensive online training and typically require retraining whenever the reward specification changes. A practical, controllable simulator should provide flexible control of behavior while preserving the realism of generated trajectories~\citep{rowe2024ctrlsim}. Existing studies also lack comparable representations of preference profiles across agent populations, which are needed for compositional-style behaviors and for adapting to variations across scenarios.

To address these limitations, we propose Controllable Neural Variational Agents (CNeVA)
, a controllable traffic simulation framework that learns from offline log-replay data. In the generated scenarios, it encodes the driving behavior of each agent by a $K$-dimensional Gaussian latent $\bm{\lambda}_{n}$ over a predetermined reward basis, such as safety, map compliance, speed, and acceleration, whose posterior is identified in closed form from each demonstration's per-channel discounted return $\bm{G}_{n}$. The inferred posterior conditions a flow-based scene generator. We propose a mixed channel-mask curriculum that exposes classifier-free guidance to arbitrary subsets of channels, allowing a user to fix some steerable axes while leaving the rest unconstrained during generation. Our main contributions in this work are as follows:
\begin{enumerate}
    \item[\textbf{C1}] We cast controllable behavior generation as variational inference over a Gaussian behavior latent with a closed-form conjugate posterior and provide a tilt-vs-regression explanatory framework that predicts which channels are identifiable from the return-shrinkage factor.
    \item[\textbf{C2}] We introduce context-residual return labeling, soft eligibility gates, and contrastive conditioning as extensions to the reward and training pipeline, producing safety controllability substantially above the hard-eligibility ablation while retaining near-ground-truth speed.
    \item[\textbf{C3}] We present a systematic accounting of the limits in our framework through the scope of drift-paired channel-steering matrices (CSMs) across three return measures, showing that map controllability is coordinate-specific and that CSM values must be interpreted jointly with physical-plausibility guardrails to avoid reward-hacking confounds.
\end{enumerate}

\section{Hierarchical Graphical Models for Simulated Agents}
\label{sec: problem}

\begin{figure}[!t]
    \begin{subfigure}[b]{.45\textwidth}
        \centering
        \begin{tikzpicture}[
            scale=0.86, transform shape,
            latent/.style={circle,fill=white,draw=black,inner sep=1pt, minimum size=20pt, font=\fontsize{10}{10}\selectfont, node distance=1},
            obs/.style={latent,fill=gray!25},
            const/.style={rectangle, inner sep=0pt, node distance=1},
            factor/.style={rectangle, fill=black,minimum size=5pt, inner sep=0pt, node distance=0.4},
            det/.style={latent, diamond},
            arrow/.style={->,=>{triangle 45}},
        ]
            \node[obs]                                  (o0)    {$\rvo_{<1}$};
            \node[latent, right=6mm of o0]              (s1)    {$\rvs_{1}$};
            \node[latent, right=6mm of s1]              (s2)    {$\rvs_{2}$};
            \node[const, right=6mm of s2]               (dots)  {$\dots$};
            \node[latent, right=6mm of dots]            (sT)    {$\rvs_{T}$};
            \node[obs, below=5mm of s1, xshift=6mm]     (o1)    {$\rvo_{1,n}$};
            \node[obs, below=5mm of s2, xshift=6mm]     (o2)    {$\rvo_{2,n}$};
            \node[obs, below=5mm of sT, xshift=6mm]     (oT)    {$\rvo_{T,n}$};
            \node[obs, above=5mm of s1]                 (m)     {$\rvm$};

            \draw[arrow]    (o0) to (s1);
            \draw[arrow]    (s1) to (o1);
            \draw[arrow]    (s1) to (s2);
            \draw[arrow]    (s2) to (o2);
            \draw[arrow]    (s2) to (dots);
            \draw[arrow]    (dots) to (sT);
            \draw[arrow]    (sT) to (oT);
            \draw[arrow]    (m) to (s1);
            \draw[arrow]    (m) to[bend left=10] (s2);
            \draw[arrow]    (m) to[bend left=20] (sT);

            \node[wrap={(o1)(o2)(oT)}]          (N-wrap)    {};
            \node[plate caption=N-wrap]         (N-caption) {$N$};
            \node[plate={(N-wrap)(N-caption)}]  (N)         {};
        \end{tikzpicture}
        \caption{Probabilistic Graphical Model for sim-agents.}
        \label{subfig: hmm}
    \end{subfigure}
    \hfill
    \begin{subfigure}[b]{.53\textwidth}
        \centering
        \begin{tikzpicture}[
            scale=0.86,transform shape,
            latent/.style={circle, fill=white, draw=black, inner sep=1pt, minimum size=20pt, font=\fontsize{10}{10}\selectfont, node distance=1},
            obs/.style={latent, fill=gray!25},
            const/.style={rectangle, inner sep=0pt, node distance=1},
            factor/.style={rectangle, fill=black, minimum size=5pt, inner sep=0pt, node distance=0.4},
            det/.style={latent, diamond},
            plate/.style={draw, rectangle,rounded corners, fit=#1},
            wrap/.style={inner sep=0pt, fit=#1},
            gate/.style={draw, rectangle, dashed, fit=#1},
            arrow/.style={->,>={triangle 45}},
            caption/.style={font=\footnotesize, node distance=0},
            plate caption/.style={caption, node distance=0, inner sep=0pt, below left=5pt and 0pt of #1.south east},
        ]
            \node[obs]                                  (o0)    {$\rvo_{<1}$};
            \node[latent, right=6mm of o0]              (s1)    {$\rvs_{1}$};
            \node[latent, right=6mm of s1]              (s2)    {$\rvs_{2}$};
            \node[const, right=6mm of s2]               (dots)  {$\dots$};
            \node[latent, right=6mm of dots]            (sT)    {$\rvs_{T}$};
            \node[obs, below=5mm of s1, xshift=6mm]     (o1)    {$\rvo_{1,n}$};
            \node[obs, below=5mm of s2, xshift=6mm]     (o2)    {$\rvo_{2,n}$};
            \node[obs, below=5mm of sT, xshift=6mm]     (oT)    {$\rvo_{T,n}$};
            \node[obs, above=5mm of s1]                 (m)     {$\rvm$};
            \node[latent, below=5mm of o2, xshift=6mm]  (lambda){$\bm{\lambda}_{n}$};

            \draw[arrow]    (o0) to (s1);
            \draw[arrow]    (s1) to (o1);
            \draw[arrow]    (s1) to (s2);
            \draw[arrow]    (s2) to (o2);
            \draw[arrow]    (s2) to (dots);
            \draw[arrow]    (dots) to (sT);
            \draw[arrow]    (sT) to (oT);
            \draw[arrow]    (m) to (s1);
            \draw[arrow]    (m) to[bend left=10] (s2);
            \draw[arrow]    (m) to[bend left=20] (sT);
            \draw[arrow]    (lambda) to (o1);
            \draw[arrow]    (lambda) to (o2);
            \draw[arrow]    (lambda) to (oT);

            \node[wrap={(o1)(o2)(oT)(lambda)}]    (N-wrap)    {};
            \node[plate caption=N-wrap]                       (N-caption) {$N$};
            \node[plate={(N-wrap)(N-caption)}]                (N)         {};
        \end{tikzpicture}
        \caption{Extended PGM with per-agent behavior latents.}
        \label{subfig: extended-hmm}
    \end{subfigure}
    \caption{Probabilistic graphical models for simulated agents. (a) The standard hidden Markov formulation from the Waymo Open Sim Agents~\citep{montali2023wosac}. (b) Our extended model augments each agent with a per-agent latent behavior profile $\bm{\lambda}_{n}$, which captures how the agent weights a shared set of reward channels. The latent profile is inferred from the observed trajectory through the reward-tilted preference factor and is later used as a controllable conditioning variable during generation.}
    \label{fig: pgm}
    \vspace{-1em}
\end{figure}
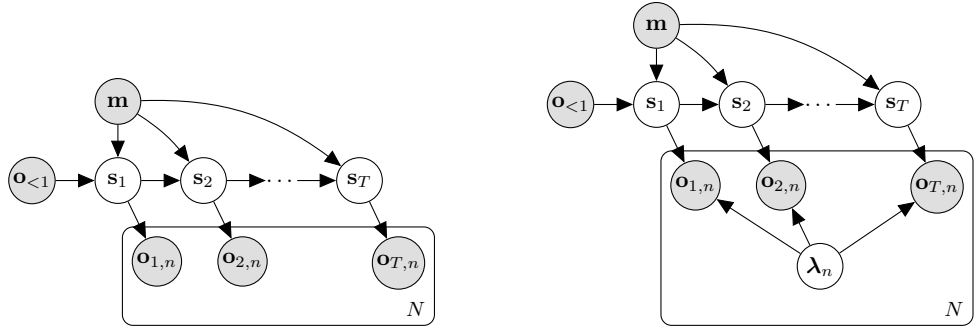

We model a multi-agent driving scenario as a Hidden-Markov process~\citep{montali2023wosac} with history $\vo_{<1}$ (\cref{subsec: exp-setup}), latent world state $\vs_{t}$, per-agent observations $\vo_{t,n}\!\in\!\R^{d}$, and time-invariant map $\vm$. \Cref{subfig: hmm} shows the resulting PGM. A model that maximizes this HMM likelihood reproduces logged behavior but provides no explicit toggle for steering generation. We therefore extend the graph (\cref{subfig: extended-hmm}) with a per-agent \emph{behavior profile} $\bm{\lambda}_{n}\!\in\!\R^{K}$, so that behavioral heterogeneity across drivers reduces to \emph{how each agent weights a shared set of reward signals}, such as safety, map compliance, speed, and acceleration. With per-step preference factor
\begin{equation}
  \psi_{t}(\vo_{t,n},\bm{\lambda}_{n})\,\triangleq\,\exp\bigl(\gamma^{t-1}\,\bm{\lambda}_{n}^{\top}\,\vr(\vo_{t,n})\bigr),
  \label{eq: psi-factor}
\end{equation}
discount factor $\gamma\!\in\!(0,1]$, and the per-channel discounted return $\bm{G}_{n}\!\triangleq\!\sum_{t=1}^{T}\gamma^{t-1}\,\vr(\vo_{t,n})\!\in\!\R^{K}$, we apply belief propagation to collapse the trajectory product into a single inner product,
\begin{equation}
  \psi(\bm{\tau}_{n},\bm{\lambda}_{n})\,\triangleq\,\prod_{t=1}^{T}\psi_{t}(\vo_{t,n},\bm{\lambda}_{n})\,=\,\exp\bigl(\bm{\lambda}_{n}^{\top}\bm{G}_{n}\bigr).
  \label{eq: cumret}
\end{equation}
Marginalizing the latent state and conditioning on history yields the relaxed joint distribution
\begin{equation}
  p(\bm{\tau},\bm{\lambda}\!\mid\!\vm)\;\propto\;\prod_{n=1}^{N}p(\bm{\lambda}_{n})\,\psi(\bm{\tau}_{n},\bm{\lambda}_{n})\,\prod_{t=1}^{T}p(\vo_{t,n}\!\mid\!\vo_{<t},\vm),
  \label{eq: simplified-generative}
\end{equation}
where $\vo_{<t}$ is a sufficient statistic for $\vs_{t}$. We defer the full HMM, the PGM, and the belief-propagation derivation to~\cref{appx-sec: posterior-derivation}. Our formulation approaches controllability by inference:
\begin{mdframed}[%
    leftline=true, linecolor=BrickRed, linewidth=2pt, topline=false,
    rightline=false, bottomline=false, backgroundcolor=dust!15
]
\noindent\emph{How can we infer the latent preference that governs the behavior of each observed agent?}
\end{mdframed}

\section{Controllable Neural Variational Agents}
\label{sec: method}

\begin{figure}[!t]
    \centering
    \includegraphics[width=\textwidth]{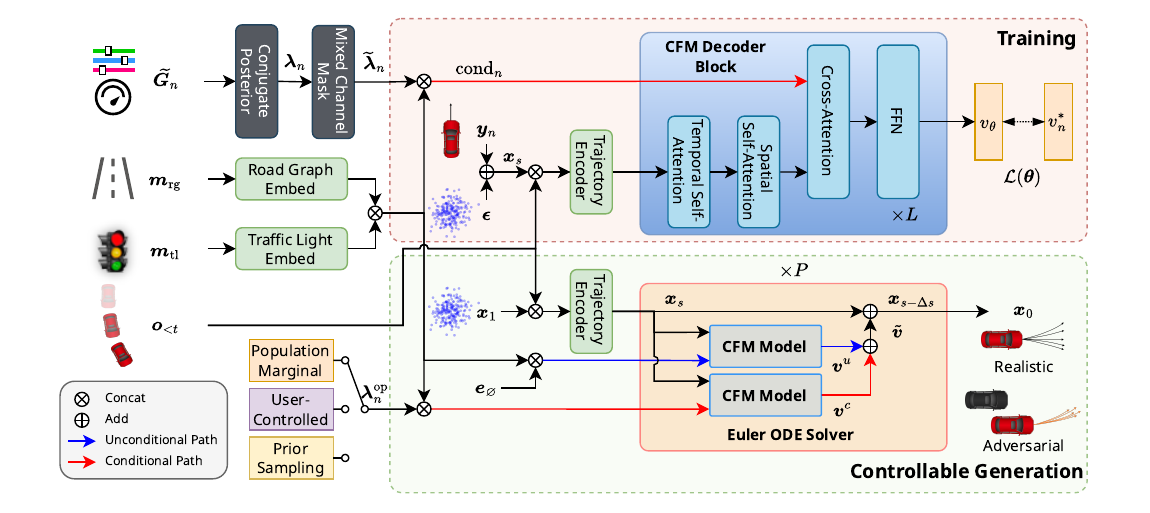}
    \caption{CNeVA infers a per-agent behavior latent $\bm{\lambda}_{n}$ from standardized reward returns via a conjugate Gaussian posterior, with mixed channel masking for classifier-free CFM training. At generation time, conditional and unconditional velocity fields are combined via CFG and an Euler ODE solver to produce population-marginal, user-controlled, and prior-sampled rollouts.}
    \label{fig: arch}
    \vspace{-1em}
\end{figure}

We propose to instantiate the factor-graph problem of~\cref{sec: problem} with a neural network through two coupled objectives: variational inference of the per-trajectory posterior over $\bm{\lambda}$ (\cref{subsec: vi}) and a conditional generator that consumes $\bm{\lambda}$ as guidance (\cref{subsec: cfm}). Sampling-time control of the trained model is described in~\cref{subsec: inference}. \Cref{fig: arch} presents our end-to-end architecture: the conjugate posterior of~\cref{subsec: vi} produces a per-agent $\bm{\lambda}_{n}$ from the standardized per-channel return, the mixed channel-mask curriculum of~\cref{subsec: cfm} trains a single conditional flow-matching velocity field that covers both conditional and unconditional inputs, and an Euler integrator with classifier-free guidance at deployment supports population-marginal, user-controlled, and prior-sampled rollouts.

\subsection{Variational Behavior Labeling}
\label{subsec: vi}

Controllable generation requires a conditional generative model $p(\bm{\tau}\!\mid\!\bm{\lambda},\vm;\vtheta)$ per agent. Because the latent $\bm{\lambda}_{n}$ is not observed in $\gD$, we treat each trajectory $\bm{\tau}_{n}$ as evidence and infer a posterior label from the per-agent factor-graph sub-problem rooted at $\bm{\lambda}_{n}$. Given the trajectory-level factor $\psi(\bm{\tau}_{n},\bm{\lambda}_{n})$ and a population prior $p(\bm{\lambda}_{n})$, variational inference~\citep{jordan1999learning} yields the per-agent approximation
\begin{equation}
  q^{\star}(\bm{\lambda}_{n})\,\gets\,\argmin_{q\in\gQ}\,\mathrm{KL}\left[q(\bm{\lambda}_{n})\,\Vert\,p(\bm{\lambda}_{n}\!\mid\!\bm{\tau}_{n})\right]\,=\,\argmax_{q\in\gQ}\E_{q}\left[\bm{\lambda}_{n}^{\top}\bm{G}_{n}\right]\,-\,\mathrm{KL}\left[q(\bm{\lambda}_{n})\,\Vert\,p(\bm{\lambda}_{n})\right].
  \label{eq: vi-objective}
\end{equation}
This objective places mass on profiles that \emph{explain} the observed returns while regularizing the inferred label toward the population prior. With a Gaussian prior, exponential-family conjugacy yields an analytic solution.

\begin{proposition}[Conjugate Gaussian posterior]
  Let $p(\bm{\lambda}_{n})\,=\,\mathcal{N}(\bm{\mu}_{0},\bm{\Sigma}_{0})$. Under~\eqref{eq: psi-factor} to \eqref{eq: vi-objective}, the variational optimum coincides with the reward-tilted posterior and is the multivariate Gaussian
  \begin{equation}
    \boxed{\quad
      q^{\star}(\bm{\lambda}_{n})\,=\,\gN\!\bigl(\bm{\mu}_{0}+\mSigma_{0}\bm{G}_{n},\,\mSigma_{0}\bigr).
    \quad}
    \label{eq: analytic-posterior}
  \end{equation}
  \label{prop: conjugate}
\end{proposition}
The posterior covariance is fixed at $\mSigma_{0}$ because the reward-tilted factor is linear in $\bm{\lambda}_{n}$, so the residual uncertainty represents population-level label noise rather than trajectory-specific epistemic uncertainty; we do not claim that $\bm{\lambda}_{n}$ recovers the agent's underlying utility in the inverse-RL sense. Derivation is in~\cref{appx-sec: posterior-derivation}. At training time we sample $\bm{\lambda}_{n}\,\sim\,q^{\star}$ with the reparameterization trick:
\begin{equation}
  \bm{\lambda}_{n}\,=\,\bm{\mu}_{0}\,+\,\mSigma_{0}\bm{G}_{n}\,+\,\mSigma_{0}^{1/2}\,\bm{\eta}_{n},\qquad\bm{\eta}_{n}\!\sim\!\gN(\bm{0},\bm{I}).
  \label{eq: reparam}
\end{equation}

\noindent\textbf{Tilt-vs-regression interpretation.} The conjugate posterior admits two equivalent readings. As a \emph{tilt}, the exponential factor $\exp(\bm{\lambda}_{n}^{\top}\bm{G}_{n})$ shifts probability mass toward profiles that explain the observed returns; as Bayesian \emph{regression}, $\bm{\lambda}_{n}$ on $\bm{G}_{n}$ is regularized toward $\bm{\mu}_{0}$ by the prior precision $\mSigma_{0}^{-1}$. Under the regression view, the return-shrinkage factor $\mSigma_{0}(\mSigma_{0}+\mSigma_{\mathrm{noise}})^{-1}$ governs how strongly each channel's return updates the posterior mean: channels with high signal-to-noise ratio (dense, trajectory-level penalties such as speed) experience strong shrinkage toward the observed return and are easy to steer, while channels with low signal-to-noise ratio (sparse, context-dependent events such as collisions) are prior-dominated and harder to control through $\bm{\lambda}$ alone. This asymmetry predicts the experimental controllability hierarchy of~\cref{subsec: exp-controllability}.

\noindent\textbf{Per-channel standardization.} The raw return $\bm{G}_{n}$ has very different per-channel scales: \cref{appx-subsec: sigma-calibration} reports empirical means $|\mu_{k}|\!\in\![3.0,\,50.1]$ and standard deviations $\sigma_{k}\!\in\![5.2,\,33.6]$. Using $\bm{G}_{n}$ verbatim in~\eqref{eq: analytic-posterior} allows one channel to dominate the conjugate update by an order of magnitude. We pre-condition with the empirical mean $\bm{\mu}_{G}$ and diagonal scale $\mathrm{diag}(\bm{\sigma}_{G})$ measured once on a fixed WOMD calibration split, and use the standardized return
\begin{equation}
  \widetilde{\bm{G}}_{n,k}\;\triangleq\;(\bm{G}_{n,k}-\bm{\mu}_{G,k})\,/\,\bm{\sigma}_{G,k},\qquad k=1,\dots,K,
  \label{eq: g-standardise}
\end{equation}
in place of $\bm{G}_{n}$ inside~\eqref{eq: analytic-posterior} and~\eqref{eq: reparam}. With $\mSigma_{0}\!=\!\bm{I}$, this brings the per-channel posterior mean to an order-unity scale, so the four channels enter the conditional generator of~\cref{subsec: cfm} on equal footing.

\subsection{Conditional Behavior Generation}
\label{subsec: cfm}

With the posterior $q^{\star}(\bm{\lambda}_{n})$ fixed, we train a posterior-labeled conditional generator against the reconstruction term of the variational lower bound,
\begin{equation}
  \vtheta^{\ast}\,\gets\,\argmax_{\vtheta\in\Theta}\E_{\bm{\tau}\sim\gD}\E_{\bm{\lambda}\sim q^{\star}(\bm{\lambda}\mid\bm{\tau})}\left[\log p(\bm{\tau}\!\mid\!\vm,\bm{\lambda};\vtheta)\right].
  \label{eq: cond-log-p}
\end{equation}
We optimize this objective via conditional flow matching~\citep{lipman2023flow}, a simulation-free surrogate that regresses a neural velocity field onto the rectified-flow target instead of evaluating the likelihood directly. We treat $p(\bm{\tau}\!\mid\!\vm,\bm{\lambda};\vtheta)$ as the implicit distribution induced by the trained velocity field. Let $\bm{y}_{n}$ denote the future motion target of agent $n$ as displacement features ($\bm{y}_{t,n}=\vo_{t,n}-\vo_{t-1,n}$). For noise $\bm{\epsilon}\sim\gN(\bm{0},\mI)$ and flow time $s\!\in\!(0,1)$ (drawn from a logit-normal distribution in our implementation), the noisy latent on the linear interpolation path and its target velocity are
\begin{equation}
  \vx_{s}\,=\,(1-s)\bm{y}_{n}+s\bm{\epsilon},\qquad\bm{v}^{\text{target}}_{n}\,\triangleq\,\bm{\epsilon}-\bm{y}_{n}.
  \label{eq: rectified-flow}
\end{equation}
A neural velocity field $\bm{v}_{\vtheta}(\vx_{s},s,\vo_{<t,n},\vm,\bm{\lambda}_{n})$ is trained to predict $\bm{v}^{\text{target}}_{n}$, transporting noisy trajectory samples back to realistic future displacements while $\bm{\lambda}_{n}$ selects among behavior modes compatible with the same scene history and map context. Architecturally, $\bm{\lambda}_{n}$ enters as an extra cross-attention token prepended to the per-agent scenario-context conditioning set.

Inference uses classifier-free guidance~\citep{ho2022cfg}. The standard binary on/off recipe is too coarse for our $K$-dimensional latent, so we expose every channel \emph{subset} during training via a four-branch curriculum (\emph{i.e., null, single-channel, two-channel, and full, with proportions $0.2/0.4/0.2/0.2$. See~\eqref{eq: mask-curriculum} in~\cref{appx-subsec: classifier-free}}). We write $\vb_{n}\!\in\!\{0,1\}^{K}$ for the resulting mask with $\vb_{n,k}\!=\!1$ marking channel $k$ as masked-out, and concatenate the value-zeroed label with the mask indicator,
\begin{equation}
  \widetilde{\bm{\lambda}}_{n}\,\triangleq\,\bigl[(\bm{1}-\vb_{n})\odot\bm{\lambda}_{n};\;\vb_{n}\bigr]\,\in\,\R^{2K},
  \label{eq: lambda-tilde}
\end{equation}
so the projection can distinguish ``channel $k$ has value $0$'' from ``channel $k$ is unobservable.'' When $\vb_{n}\!=\!\bm{1}_{K}$, the $\bm{\lambda}$ token is overridden by a learned null embedding $\ve_{\varnothing}\!\in\!\R^{d_{h}}$, recovering the unconditional path required for the CFG combination at inference. The full training loss is
\begin{equation}
  \boxed{\quad
  \gL(\vtheta)\,=\,\sum_{n=1}^{N}\,\mathbb{E}_{q^{\star}(\bm{\lambda}_{n}),\,\vb_{n},\,s,\,\bm{\epsilon}}\!\left[\,\bigl\Vert\bm{v}_{\vtheta}(\vx_{s},s,\vo_{<t,n},\vm,\widetilde{\bm{\lambda}}_{n})\,-\,\bm{v}^{\text{target}}_{n}\bigr\Vert_{2}^{2}\,\right].
  \quad}
  \label{eq: cfm-cfg-loss}
\end{equation}
By design, the null branch, active when $\vb_{n}\!=\!\bm{1}_{K}$, learns the population path $p(\bm{\tau}\!\mid\!\vm)$ on the training set. In contrast, the partially- and fully-conditional branches learn the steered path $p(\bm{\tau}\!\mid\!\vm,\bm{\lambda})$. Architectural details are given in~\cref{appx-sec: model-details}.

\subsection{Inference and Controllable Generation}
\label{subsec: inference}

At deployment, the operator supplies a per-agent $\bm{\lambda}_{n}^{\mathrm{op}}$ in the standardized coordinates of~\eqref{eq: g-standardise}; the trajectory is drawn from $p_{\vtheta}(\bm{\tau}_{n}\!\mid\!\vm,\bm{\lambda}_{n}^{\mathrm{op}})$ via the rectified-flow Euler sampler with classifier-free guidance~\citep{ho2022cfg}:
\begin{equation}
  \widetilde{\bm{v}}_{\vtheta}^{\,w}(\vx_{s},\bm{\lambda}_{n}^{\mathrm{op}})\,\triangleq\,(1\!+\!w)\,\bm{v}_{\vtheta}(\vx_{s},s,\vo_{<t,n},\vm,\bm{\lambda}_{n}^{\mathrm{op}})\,-\,w\,\bm{v}_{\vtheta}(\vx_{s},s,\vo_{<t,n},\vm,\ve_{\varnothing}),
  \label{eq: cfg-velocity}
\end{equation}
with guidance scale $w\!\geq\!0$. The population-marginal, user-controlled, and prior-sampled operator regimes, the receding-horizon patch scheme, and the full sampler are detailed in~\cref{appx-subsec: euler-sampler}.

\subsection{Return Labeling and Eligibility}
\label{subsec: context-residual}

\paragraph{Context-residual returns.} The raw return $G_{n,k}$ conflates driving style with scenario difficulty. For example, a highway scene can yield low off-road penalties regardless of the individual driving style. Therefore, we residualize each per-channel return against a scene-context baseline, $G^{\mathrm{cr}}_{n,k}\!=\!G_{n,k}-\bar{G}_{k}(\vm_{n})$, isolating the behavioral signal. We use this as the primary return measure for all CNeVA training and evaluation (\cref{appx-subsec: context-residual-detail}). As a coordinate-specific alternative used only in the map conformity ablation in~\cref{appx-subsec: map-reward-ablation}, we also define a \emph{lane-centerline return}\label{subsec: lane-centerline} from the signed lateral offset to the nearest lane centerline (\cref{appx-subsec: lane-centerline-detail}).

\paragraph{Soft eligibility gates.}\label{subsec: soft-eligibility} The safety and map conformity channels use per-agent eligibility gates that decide which agents receive non-trivial labels. The hard-gated baseline labels only agents within fixed clearance, time-to-collision (TTC), or road-margin thresholds, introducing a discontinuity at the threshold and excluding the majority of safe, on-road agents from supervision. We replace these binary gates with a \emph{smooth exponential decay} in the per-step pairwise risk, so that all valid agents receive labels. In contrast, agents far from hazards contribute negligibly. The design preserves gradient signal for near-threshold agents and fixes the safety-CSM erosion documented in~\cref{subsec: exp-plausibility}. We defer details about the decay formulas and scales ($\tau_{c},\tau_{t},\tau_{m}$) to~\cref{appx-subsec: soft-eligibility-detail}.

\paragraph{Contrastive conditioning.}\label{subsec: contrastive-cond} We further augment training with a paired steered-null forward pass over the same flow-time noise, encouraging the velocity field to learn the steered-versus-unsteered \emph{difference} and improving robustness under closed-loop drift (\cref{appx-subsec: contrastive-detail}).

  \section{Related Work}
\label{sec: related}

\paragraph{Data-Driven Traffic Simulation for AV Testing.}
Unlike open-loop trajectory forecasting~\citep{nayakanti2023wayformer,shi2024mtrpp}, scenario generation requires agents that react to AV actions, whether by modeling the joint trajectory distribution~\citep{ngiam2022scenetransformer} or via a second-stage refinement~\citep{zhang2022trajgen}. Tokenized models that discretize actions~\citep{seff2023motionlm,philion2024trajeglish,wu2024smart,zhang2025catk,chang2026spacer,zhou2024behaviorgpt} or spaces~\citep{hu2024gump,zhou2024behaviorgpt}, and that address covariate shift via rollout during training~\citep{zhang2025catk}, improve the behavioral realism, but these autoregressive approaches can lose behavioral diversity to dominant motion patterns. Generative VAE~\citep{suo2021trafficsim,zhang2023trafficbots,liu2026stage}, diffusion~\citep{jiang2023motiondiffuser,cao2024rlhf,jiang2024scenediffuser,tan2025scenediffuser2}, and flow-matching~\citep{lipman2023flow,xing2025goalflow,tan2025flowplanner} models capture multi-modal behavior well but expose little interpretability of driving style, and self-play RL often sacrifices realism or needs reference models~\citep{cornelisse2024hrppo,cusumano2025gigaflow} for regularization~\citep{chang2026spacer}. CNeVA instead infers a closed-form posterior behavioral latent over an explicit reward basis, which acts as a single guidance token for a rectified-flow generator, improving the realism and diversity of controllable generation.

\paragraph{Controllable Agent Behavior Generation.}
Modeling driving style as a controllable parameter enables behavior-specific generation~\citep{yin2021routegan,rempe2022accidentprone,chang2023socialedit,liu2026stage}. In generative models, controllability is typically injected through guidance (hard constraints~\citep{jiang2024scenediffuser,tan2025scenediffuser2}, hand-crafted differentiable costs~\citep{huang2024vbd,chang2024safesim}, STL rules~\citep{zhong2023ctg}, conditional priors~\citep{Pronovost2023scenariodiffusion}, or LLM-prompted attractors~\citep{zhong2023ctgpp}), but each typically requires redesigning the guidance mechanism for new behaviors. Driving style can also be read as explicit or implicit reward preferences~\citep{rowe2024ctrlsim,cusumano2025gigaflow,rowe2025scenariodreamer}, though per-channel rewards vary across scenarios. CNeVA instead learns the latent behavioral profile directly, keeping inference cheap and the editable interface compact while exposing per-channel controllability handles that map cleanly to interpretable driving-style axes.

  \section{Experiments}
\label{sec: experiment}

We evaluate CNeVA on the Waymo Open Motion Dataset~\citep{ettinger2021womd} with four questions:
\begin{enumerate}
    \item[\textbf{RQ1}] Does the unconditional path generate realistic behavior (\cref{subsec: exp-fidelity})?
    \item[\textbf{RQ2}] Does the learned latent steer per-channel returns with physical plausibility (\cref{subsec: exp-controllability})?
    \item[\textbf{RQ3}] What do ablations without soft eligibility reveal about safety controllability erosion and reward-hacking confounds (\cref{subsec: exp-plausibility})?
\end{enumerate}
In addition, we investigate how sensitive map controllability is to the choice of return measure and defer our discussion and results to~\cref{appx-subsec: map-reward-ablation}.

\subsection{Experiment Setup}
\label{subsec: exp-setup}

\noindent We evaluate on WOMD~\citep{ettinger2021womd} under the WOSAC protocol~\citep{montali2023wosac}: given $1.1$\,s of $10$\,Hz history, the model generates $8$\,s rollout for each scenario, with up to $N\!=\!128$ agents, $\vo_{t,n}\!\in\!\R^{9}$ (pose, heading, planar velocity, bounding box). We report open-loop fidelity as $\mathrm{minADE}_{S}$ and the WOSAC composite meta-metric. We evaluate controllability by the channel steering matrix (CSM) and the trajectory divergence $D(\bm{e}_{a}, \bm{e}_{b})$ (\cref{subsec: exp-controllability}). We build the proposed CNeVA with $d_{h}\!=\!512$, six Transformer decoder layers with eight attention heads, and $K\!=\!4$ reward channels. Each channel is a negative sum of penalties. Hence, a higher $G_{n,k}$ denotes less of the penalized quantity: \emph{safety} (collision count, proximity to nearest object, inverse time-to-collision), \emph{map} (offroad indicator, road-edge distance, computed under the context-residual formulation of~\cref{subsec: context-residual}), \emph{speed} (linear and angular velocity magnitudes), \emph{accel} (linear and angular acceleration magnitudes). We use standard normal behavior prior $(\bm{\mu}_{0},\mSigma_{0})\!=\!(\bm{0},\bm{I})$. Returns are standardized by the training dataset statistics (\cref{appx-subsec: sigma-calibration}). The CFM head predicts a $6$-d motion target, with bounding box length, width, and height agent-static and reused from history. During sampling, we apply the Euler integrator with $10$ steps, and the receding-horizon patch length is $\ell\!=\!16$. The main model uses soft eligibility gates (\cref{subsec: soft-eligibility}) with decay scales $\tau_{c}\!=\!2.0$\,m, $\tau_{t}\!=\!3.0$\,s, $\tau_{m}\!=\!1.0$\,m, future-query map conditioning with $k\!=\!16$ nearest neighbors, and random mid-step anchor sampling. All main results are evaluated at $200,000$ training steps.

\subsection{Benchmark}
\label{subsec: exp-fidelity}


\begin{table*}[!t]
    \caption{\textbf{Waymo Open Sim-Agents Challenge Realism metrics on the \emph{test} split of WOMD.} We rank the baseline methods by realism on the 2025 challenge leaderboard. \emph{Oracle} shows the results evaluated with the ground truth. Methods with full or partial controllability are marked in \textcolor[HTML]{8e6f3e}{gold}.}
    \centering
    \small
    \setlength{\tabcolsep}{6pt}
    \begin{tabular}{@{}l|ccc|cc@{}}
    \toprule
    \textbf{Model} & \textbf{Kinematic ($\uparrow$)} & \textbf{Interactive ($\uparrow$)} & \textbf{Map-based ($\uparrow$)} & \textbf{Realism ($\uparrow$)} & \textbf{minADE ($\downarrow$)} \\
    \midrule
    Constant Velocity   & $0.2253$  & $0.4327$  & $0.4535$  & $0.3985$  & $7.5148$  \\
    TrafficBots~V1.5~\citep{zhang2023trafficbots}   & $0.4304$  & $0.7114$  & $0.8360$  & $0.6988$    & $1.8825$    \\
    SceneDiffuser~\citep{jiang2024scenediffuser}   & $0.4295$   & $0.7681$  &  $0.7756$  & $0.7030$ & $1.7670$  \\
    \textcolor[HTML]{8e6f3e}{VBD}~\citep{huang2024vbd}  & $0.4169$  & $0.7819$  & $0.8137$ & $0.7200$ & $1.4743$ \\
    TrajTok~\citep{trajtok2025} &   $0.4887$    & $0.8116$  & $0.9207$  & $0.7852$  & $1.3179$  \\
    SMART-R1~\citep{pei2026advancingmultiagenttrafficsimulation}   & $0.4940$  & $0.8109$  & $0.9194$  & $0.7855$  & $1.2990$	\\
    Oracle~\citep{jiang2024scenediffuser} & $0.5565$    & $0.8576$  & $0.9593$  & $0.8330$  & $0.0000$ \\
    \midrule
    \textcolor[HTML]{8e6f3e}{\textbf{CNeVA (Ours)}} & $0.4732$  & $0.7482$  & $0.8091$  & $0.7145$  & $1.8029$\\
    \bottomrule
    \end{tabular}
    \label{tab: benchmark}
    \vspace{-1em}
\end{table*}

\Cref{tab: benchmark} positions CNeVA against the full WOSAC realism spectrum. On the official 2025 test server, our model attains a Realism meta-metric of $0.7145$ with $\mathrm{minADE}\!=\!1.80$\,m, using inference-time guidance and best-of-128 selection on the frozen model. The benchmark result shows that CNeVA can achieve competitive behavioral realism. We notice that the top of the leaderboard is dominated primarily by non-controllable closed-loop and tokenized imitation models, and our remaining gap to them is concentrated in the collision and off-road components, consistent with open-loop error accumulation over the $8$\,s rollout. Crucially, the guidance and selection are post-hoc operations on the frozen model and leave the controllability results of \cref{subsec: exp-controllability} unchanged, as they are measured on the same base model. Our CNeVA achieves this realism without closed-loop fine-tuning or autoregressive token-prediction objectives, while exposing a per-channel, reward-conditioned controllability interface that higher-realism imitation models lack.

\subsection{Controllability}
\label{subsec: exp-controllability}

To quantify controllability, we compare steered CNeVA with the unconditional baseline using drift-paired channel-steering matrices (CSMs) on the WOMD validation split. For each channel $k$ we feed $\bm{\lambda}_{n}^{\mathrm{op}}\!=\!\rho\bm{e}_{k}$ with the single-channel mask $\vb_{n,j}\!=\!\mathbb{I}\{j\!\neq\!k\}$ and compare against the null path under the same flow-time random seed (\emph{drift-paired evaluation}). The CSM diagonal $\Delta R_{k}\!=\!G^{\mathrm{steered}}_{k,k}-G^{\mathrm{base}}_{k}$ measures per-channel response. All results use $\rho\!=\!1$, $w\!=\!1.5$, and context-residual returns unless stated otherwise.

\begin{table}[t]
  \centering
  \small
  \setlength{\tabcolsep}{4pt}
  \caption{Channel steering matrix (CSM) diagonals ($\Delta R_k$) for CNeVA with soft eligibility gates, evaluated at 200K training steps. We report results with open-loop sampling and context-residual return at $\rho\!=\!1$. The $w\!=\!0$ column is latent-only (classifier-free guidance off); the $w\!=\!1.5$ column is latent with classifier-free guidance.}
  \label{tab: csm-main}
  \begin{tabular}{lccc}
    \toprule
    & \multicolumn{3}{c}{\textbf{Open-loop}} \\
    \cmidrule(lr){2-4}
    \textbf{Channel} & Uncond. & CNeVA ($w\!=\!0$) & CNeVA ($w\!=\!1.5$) \\
    \midrule
    Safety & $+0.06_{\textcolor{darkgray}{\pm 0.00}}$  & $+0.29_{\textcolor{darkgray}{\pm 0.00}}$ & $+0.66_{\textcolor{darkgray}{\pm 0.10}}$ \\
    Map    & $+0.06_{\textcolor{darkgray}{\pm 0.00}}$  & $+0.24_{\textcolor{darkgray}{\pm 0.00}}$ & $+0.61_{\textcolor{darkgray}{\pm 0.14}}$ \\
    Speed  & $-3.33_{\textcolor{darkgray}{\pm 0.00}}$  & $+3.21_{\textcolor{darkgray}{\pm 0.00}}$ & $+8.15_{\textcolor{darkgray}{\pm 0.07}}$ \\
    Accel  & $+4.77_{\textcolor{darkgray}{\pm 0.00}}$  & $+4.19_{\textcolor{darkgray}{\pm 0.00}}$ & $+8.76_{\textcolor{darkgray}{\pm 0.07}}$ \\
    \midrule
    Null ADE & \multicolumn{3}{c}{$1.113_{\textcolor{darkgray}{\pm 0.011}}$} \\
    Offroad  & \multicolumn{3}{c}{$32.5_{\textcolor{darkgray}{\pm 0.2}}\%$} \\
    \bottomrule
  \end{tabular}
  \vspace{-1em}
\end{table}

\paragraph{Per-channel response.}
\Cref{tab: csm-main} summarizes the evaluated four-channel CSM. \Cref{fig: controllability-overview} in the appendix further contrasts these diagonals against the hard-eligibility ablation. The unconditional baseline produces near-zero diagonals on all channels, confirming that the null path does not steer by construction. CNeVA shows a clear hierarchy: the dense speed and acceleration channels respond strongest, followed by safety and map conformity, consistent with the tilt-vs-regression prediction in~\cref{subsec: vi} that dense kinematic penalties produce high signal-to-noise returns while sparse semantic penalties are prior-dominated. The $w\!=\!0$ (latent-only) column isolates what the latent encodes without guidance. Specifically, the kinematic channels are already strongly steerable, while the sparse safety and map channels roughly double under $w\!=\!1.5$. High-SNR channels are identifiable from the latent alone; sparse channels rely on classifier-free guidance. Results indicate that all four channels steer physically validly (\cref{subsec: exp-plausibility}).

\paragraph{Fidelity.}
CNeVA achieves a null-path $\mathrm{minADE}$ of $1.113 \pm 0.011$\,m and an offroad rate of $32.5 \pm 0.2\%$ at 200K steps on the WOMD validation split and $\mathrm{minADE}\!=\!1.80$\,m on the WOMD testing split. The off-road rate climbs over the rollout to roughly $1.3$ to $1.9\times$ the logged-data rate by $8$\,s (logged ${\approx}\,17\%$), reflecting open-loop drift rather than a static gap. Under the identical local evaluation protocol, the hard-eligibility ablation reaches $\mathrm{minADE}\!=\!1.112 \pm 0.015$\,m (\cref{tab: plausibility}), confirming that the soft eligibility gates do not degrade trajectory fidelity.

\paragraph{Operating regime.}
All main results use the calibrated point $\rho\!=\!1$, $w\!=\!1.5$. At $\rho\!=\!5$ all four channels stay monotonically positive with no sign-flip ($\Delta R_{\mathrm{speed}}\!=\!{+}14.10$, $\Delta R_{\mathrm{accel}}\!=\!{+}11.73$, $\Delta R_{\mathrm{safety}}\!=\!{+}0.85$, $\Delta R_{\mathrm{map}}\!=\!{+}0.42$; 5-seed means), but guardrails degrade (speed-steered stall ${+}2.1\!\to\!{+}4.6$\,pp, retained speed $94.7\%\!\to\!89.1\%$), so we report at the calibrated point.

\begin{figure}[t]
  \centering
  \begin{subfigure}[t]{0.30\linewidth}\centering
    \includegraphics[width=\linewidth]{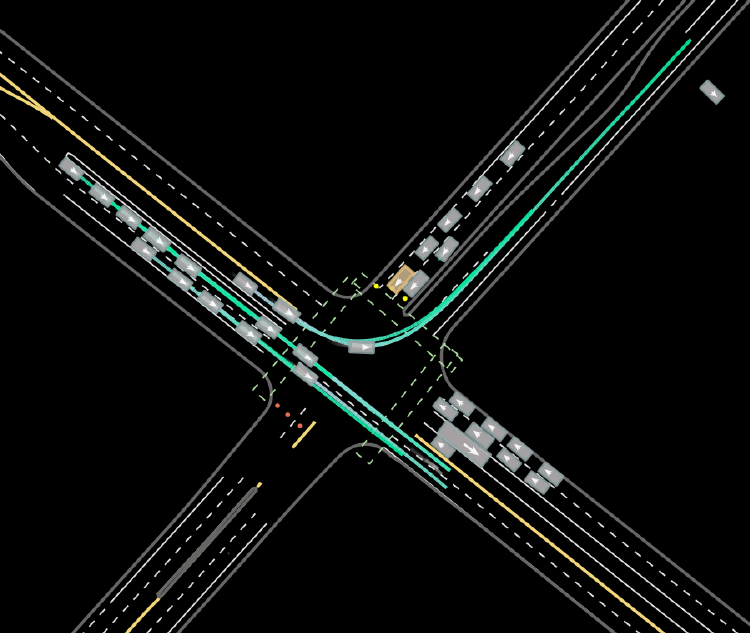}%
    \caption{Ground Truth}\label{fig: e40k-gt}
  \end{subfigure}\hfill
  \begin{subfigure}[t]{0.30\linewidth}\centering
    \includegraphics[width=\linewidth]{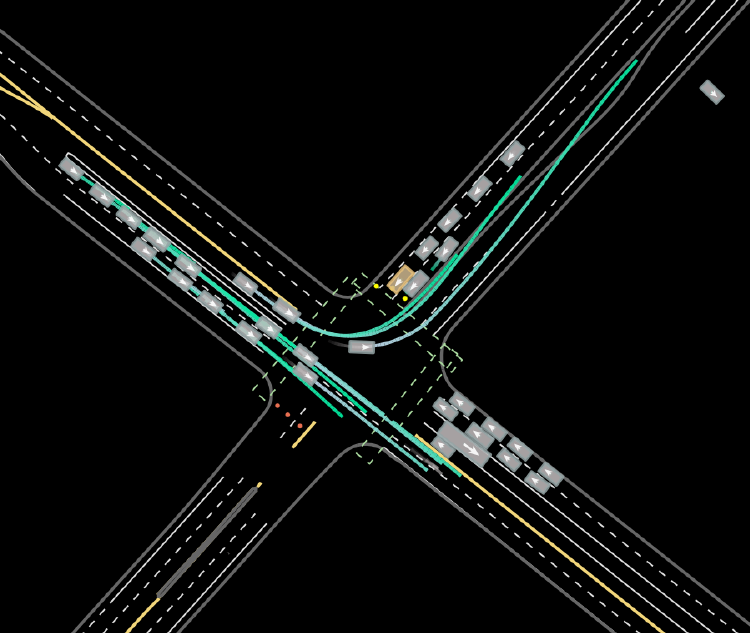}%
    \caption{Unconditional ($\bm{\lambda}\!=\!\bm{0}$)}\label{fig: e40k-null}
  \end{subfigure}\hfill
  \begin{subfigure}[t]{0.30\linewidth}\centering
    \includegraphics[width=\linewidth]{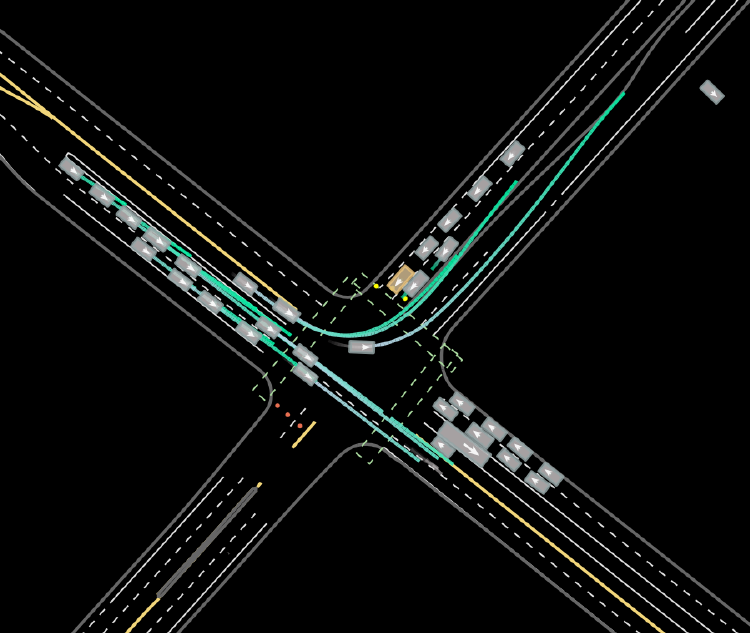}%
    \caption{Safety-Dominant}\label{fig: e40k-safety}
  \end{subfigure}\\[0.3em]
  \begin{subfigure}[t]{0.30\linewidth}\centering
    \includegraphics[width=\linewidth]{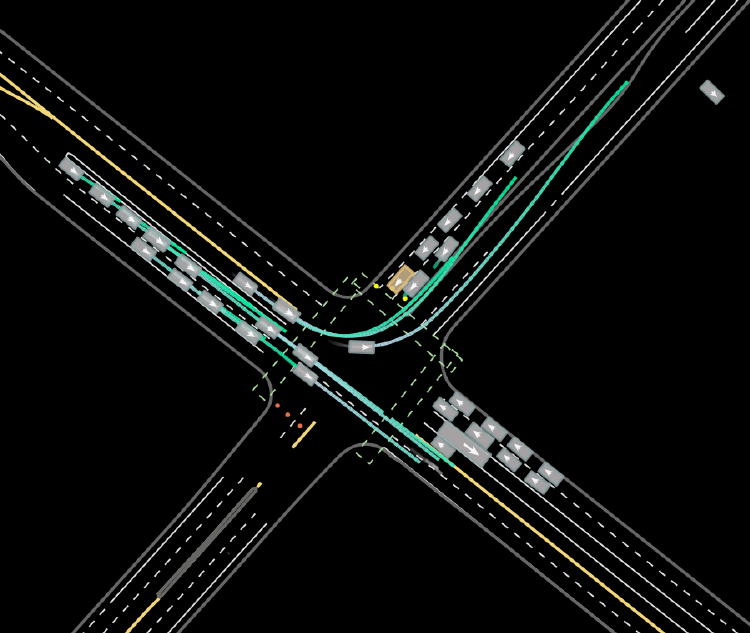}%
    \caption{Map-Dominant}\label{fig: e40k-map}
  \end{subfigure}\hfill
  \begin{subfigure}[t]{0.30\linewidth}\centering
    \includegraphics[width=\linewidth]{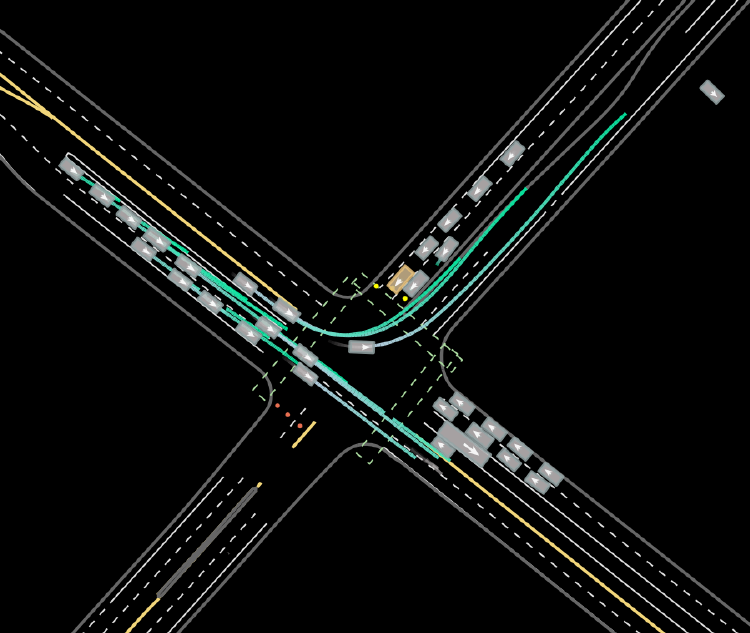}%
    \caption{Speed-Dominant}\label{fig: e40k-speed}
  \end{subfigure}\hfill
  \begin{subfigure}[t]{0.30\linewidth}\centering
    \includegraphics[width=\linewidth]{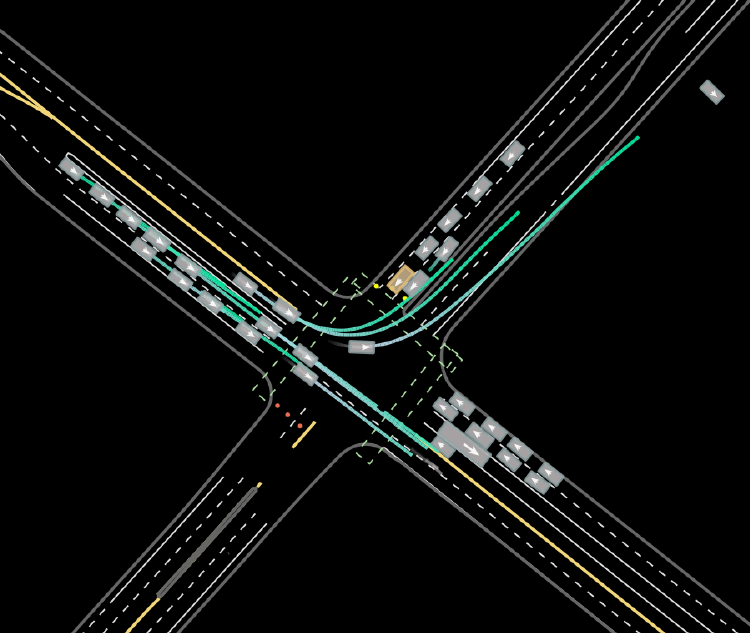}%
    \caption{Accel-Dominant}\label{fig: e40k-accel}
  \end{subfigure}
  \caption{CNeVA qualitative rollouts at $\rho\!=\!1$, $w\!=\!1.5$.}
  \label{fig: qualitative-e40k}
  \vspace{-1em}
\end{figure}

\noindent\textbf{Qualitative results.}
\Cref{fig: qualitative-e40k} shows rollouts under the four one-hot preferences against the ground-truth future and the null path. Safety and speed steering produce visible shifts matching the positive CSM diagonals. Acceleration yields smoother turning arcs, and the map panel is nearly indistinguishable from the null path (cf.~\cref{appx-subsec: map-reward-ablation}).

\subsection{Eligibility Gating and Reward Hacking}
\label{subsec: exp-plausibility}

The CSM diagonals in~\cref{tab: csm-main} measure return-space response but fail to verify whether the steered trajectories remain physically feasible. For instance, a model that maximizes the speed return by suppressing agent motion (stalling) achieves a large $\Delta R_{\mathrm{speed}}$ without producing useful behavior. We therefore complement the CSM with physical plausibility diagnostics: the stall fraction (proportion of agents with per-step displacement $<0.1$\,m), the mean speed of steered agents as a fraction of logged ground-truth speed, and the offroad rate at the $8$\,s horizon. We evaluate two ablation checkpoints alongside the main model to illustrate two failure modes that soft eligibility addresses.

\begin{figure}[t]
  \centering
  \includegraphics[width=\linewidth]{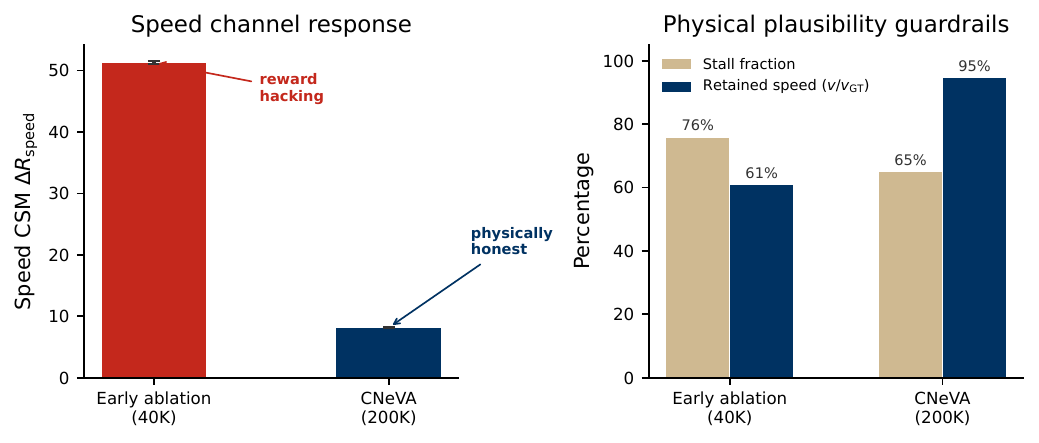}
  \caption{Reward-hacking contrast. \emph{Left}: speed CSM is inflated $6\times$ in the early checkpoint relative to the main model. \emph{Right}: physical plausibility guardrails reveal that the early ablation achieves its CSM by stalling (76\% stall, 61\% of GT speed), while the main model retains 95\% of GT speed.}
  \label{fig: reward-hacking-contrast}
  \vspace{-1em}
\end{figure}

\paragraph{Reward hacking.}
\Cref{fig: reward-hacking-contrast} (with \cref{tab: plausibility}) reveals that the early-stage ablation's speed CSM is inflated by reward hacking: it maximizes the speed return by suppressing motion (most steered agents stall) rather than by maintaining realistic velocities. CSM values must therefore be read jointly with the physical-plausibility guardrails, under which the speed response is physically valid.

\paragraph{Safety erosion under hard eligibility.}
The hard-eligibility ablation adds future-query map conditioning and random mid-step anchoring while retaining hard binary gates. Its safety CSM drops to ${+}0.21 \pm 0.22$, which is statistically indistinguishable from zero. Under hard gates, most agents (clearance $>5$\,m, TTC $>6$\,s) receive \emph{no safety label}, so the generator sees a near-zero safety signal. Replacing the thresholds with exponential decay (\cref{subsec: soft-eligibility}) preserves the signal for near-threshold agents and yields ${+}0.66 \pm 0.10$. The two values are measured across different eligibility populations and are not directly comparable in absolute terms; the relevant observation is that soft eligibility yields a statistically significant positive response, whereas hard eligibility does not.

\paragraph{Physical validity of safety steering.}
Under safety-channel steering, the main model's stall fraction increases by only ${+}0.9$ percentage points relative to the unconditional baseline, and steered agents retain $97.9\%$ of ground-truth speed. The speed reduction is consistent across all four channels, indicating that the context-residual labeling successfully decorrelates safety from speed. The safety CSM of ${+}0.66$ reflects genuine defensive-driving behavior (spacing, yielding) rather than a slow-equals-safe confound.

  \section{Conclusion}

We presented CNeVA, a controllable offline sim-agent framework that represents each traffic participant with a per-agent Gaussian behavior latent inferred in closed form from per-channel discounted returns. CNeVA conditions a mixed channel-mask flow-matching generator, supporting classifier-free guidance under null, partial, or fully specified operator preferences. Soft eligibility gates replace hard binary thresholds with smooth exponential decay, preserving gradient signal for near-threshold agents. Experiments on WOMD show that CNeVA reaches a WOSAC test-server realism of $0.71$ (mid-spectrum; kinematic realism comparable to the leading imitation models, with collision/off-road and minADE limited by open-loop drift) while producing physically valid controllability: speed and acceleration respond monotonically without the stall-induced reward hacking observed in an early ablation, and safety steering is substantially above the hard-eligibility ablation while retaining near-ground-truth speed.

\noindent\textbf{Limitations.} Map controllability is coordinate-specific: the context-residual return produces a positive response (${+}0.61$), but the physical-offroad ($-0.12$) and lane-centerline (${\approx}\,0$) measures fail to produce controllable steering (\cref{appx-subsec: map-reward-ablation}). This sensitivity to the choice of reward definition exposes a structural limitation of the current framework. All four channels remain monotonically positive through $\rho\!=\!5$; however, physical guardrails degrade at larger magnitudes. Therefore, the main results are reported at the calibrated operating point.

\noindent\textbf{Future work.} Richer opportunity-aware reward decompositions that separate spatial from temporal map compliance, closed-loop training with contrastive objectives, and stronger latent-conditioning mechanisms beyond one-hot steering are natural next steps.

  \section*{Acknowledgments}
\label{sec: ack}
The authors appreciate the Google TPU Research Cloud (TRC) for supporting access to TPUs.

  \putbib[references]
\end{bibunit}

\begin{bibunit}[plainnat]
  \newpage
\appendix
\begin{appendices}
    \startcontents[appendices]

    \printcontents[appendices]{l}{1}{\setcounter{tocdepth}{2}}
    \newpage

    \section{Notations}
    \label{appx-sec: notations}

    For reference, \cref{tab: notation} lists all notation used in this paper.
    \begin{table}[!htbp]
    \centering
    \caption{Table of Notations}
    \begin{tabular}{@{}l|l@{}}
        \toprule
        \textbf{Notation}               & \textbf{Description}              \\
        \midrule
        \multicolumn{2}{@{}l}{\emph{Scalars}} \\
        $d$ & Dimension of motion state observations. \\
        $N$ & Number of agents in a scenario. \\
        $T$ & Total number of time steps in a scenario. \\
        $T_{f}$ & Number of future time steps predicted per CFM rollout. \\
        $K$ & Number of reward channels in the semantic decomposition. \\
        $\gamma$ & Discount factor for the per-channel return; $\gamma\in(0,1]$. \\
        $s$ & Rectified-flow time; $s\in[0,1]$. \\        \midrule
        \multicolumn{2}{@{}l}{\emph{Hidden-Markov Process and Sim-Agent Scenario}} \\
        $\rvs_{t}$ & Latent world state at time step $t$ (random variable). \\
        $\rvo_{t,n}$ & Observation of the $n$-th agent at time step $t$ (random variable). \\
        $\rvm$ & Static map context (random variable; observed). \\        $\vo_{t},\,\vo_{t,n}$ & Observation / observation of the $n$-th agent at time $t$ (vector value). \\
        $\vs_{t}$ & Latent state at time $t$ (vector value). \\
        $\vm$ & Static map context (vector value). \\
        $\bm{\tau}$ & Trajectory sequence $[\vo_{1},\dots,\vo_{T}]$. \\
        $\gS,\,\gO$ & Latent state / observation space. \\
        $\gH$ & Hidden-Markov Process tuple over latent state $\gS$ and observation space $\gO$. \\
        $\gD$ & Training dataset of logged scenarios. \\
        \midrule
        \multicolumn{2}{@{}l}{\emph{Reward and Behavior Latent}} \\
        $\vr(\vo_{t,n})$ & $K$-channel per-step reward decomposition. \\
        $\bm{G}_{n}$ & Per-channel discounted observed return of agent $n$. \\
        $\vb_{n}$ & Per-agent channel mask, $\vb_{n}\!\in\!\{0,1\}^{K}$; $\vb_{n,j}\!=\!1$ marks channel $j$ as unobservable. \\
        $\bm{\lambda}_{n}$ & Per-agent behavior profile latent; $\bm{\lambda}_{n}\in\R^{K}$. \\
        $\bm{\mu}_{0}$ & Mean of the Gaussian prior on $\bm{\lambda}_{n}$. \\
        $\mSigma_{0}$ & Covariance of the Gaussian prior on $\bm{\lambda}_{n}$. \\
        $\sigma_{0,k}^{2}$ & Per-channel prior variance ($k$-th diagonal entry of $\mSigma_{0}$). \\
        $q^{\star}(\bm{\lambda}_{n})$ & Analytic conjugate Gaussian posterior. \\
        $\bm{\mu}^{\star}_{n}$ & Posterior mean $\bm{\mu}_{0}+\mSigma_{0}\bm{G}_{n}$. \\
        $\tau_{c}$ & Clearance decay scale for the safety soft eligibility gate (metres). \\
        $\tau_{t}$ & Time-to-collision decay scale for the safety soft eligibility gate (seconds). \\
        $\tau_{m}$ & Margin decay scale for the map soft eligibility gate (metres). \\
        \midrule
        \multicolumn{2}{@{}l}{\emph{Posterior Sampling}} \\
        $\bm{\eta}_{n}$ & Standard-normal noise used in the reparameterized sample. \\
        \midrule
        \multicolumn{2}{@{}l}{\emph{$\bm{\lambda}$-Conditional CFM Decoder}} \\
        $\bm{v}_{\vtheta}$ & $\bm{\lambda}$-conditional rectified-flow velocity field. \\
        $\vx_{s}$ & Noisy rectified-flow latent at time $s$. \\
        $\bm{y}_{n}$ & Clean future displacement target for agent $n$, $\R^{T_{f}\times d}$. \\
        $\bm{v}^{\star}_{n}$ & Rectified-flow velocity target $\bm{\epsilon}-\bm{y}_{n}$. \\
        $\bm{\epsilon}$ & Standard-normal noise sample for the rectified-flow forward process. \\
        $d_{h}$ & Hidden dimension of the scene-decoder backbone. \\
        $\vtheta$ & Parameters of the $\bm{\lambda}$-conditional CFM decoder. \\
        $\Delta s$ & Euler integrator step in flow time. \\
        \midrule
        \multicolumn{2}{@{}l}{\emph{Probability and Loss}} \\
        $\gN(\bm{\mu},\mSigma)$ & Multivariate Gaussian distribution with mean $\bm{\mu}$ and covariance $\mSigma$. \\
        $\gU(0,1)$ & Uniform distribution on $[0,1]$. \\
        $\E$ & Expectation operator. \\
        $\KL[\,\cdot\,\Vert\,\cdot\,]$ & Kullback--Leibler divergence between two distributions. \\
        $\gL(\bm{\theta})$ & Classifier-free CFM training loss. \\
        \bottomrule
    \end{tabular}
    \label{tab: notation}
\end{table}

    \section{Derivation of the Conjugate Gaussian Posterior}
    \label{appx-sec: posterior-derivation}

    This appendix derives the analytic posterior of~\Cref{prop: conjugate} from two perspectives. \Cref{appx-subsec: posterior-completing-square} gives the textbook completing-the-square argument used in the main text. \Cref{appx-subsec: posterior-bp} re-derives the same posterior via sum-product belief propagation on the PGM in~\Cref{subfig: extended-hmm}, which makes the conditional-independence assumptions and the role of the discount factor $\gamma$ explicit.
    Variational identity. The objective in Eq.~(4) is the KL projection onto the reward-tilted posterior: for any density $q$, the KL to the tilted posterior equals the KL to the Gaussian prior minus $\mathbb{E}_{q}[\lambda_n^T G_n]$, plus a log-normalizer independent of $q$. For a positive-definite Gaussian prior, the log-normalizer is finite and equal to $\mu_0^T G_n + \tfrac{1}{2}G_n^T\Sigma_0G_n$. Thus, the maximizer in Eq.~(4) is the same distribution recovered by completing the square below.
    Interpretation. Since the reward-tilted factor is linear in lambda, the conjugate update shifts the posterior mean but leaves the covariance fixed at the prior covariance. The residual uncertainty should therefore be read as population-level label noise, not trajectory-specific epistemic uncertainty; CNeVA uses lambda as a behavior-profile coordinate induced by the chosen reward basis, not as an identified inverse-RL utility.

    \subsection{Completing the Square}
    \label{appx-subsec: posterior-completing-square}

    Combining the Gaussian prior $p_{0}(\bm{\lambda}_{n})=\gN(\bm{\mu}_{0},\mSigma_{0})$ with the trajectory-level un-normalized factor $\psi(\bm{\tau}_{n},\bm{\lambda}_{n})=\exp(\bm{\lambda}_{n}^{\top}\bm{G}_{n})$ from~\eqref{eq: cumret}, the un-normalized log-posterior writes
    \begin{equation}
      \log p(\bm{\lambda}_{n}\!\mid\!\vo_{1:T,n})\,=\,-\tfrac{1}{2}\bigl(\bm{\lambda}_{n}-\bm{\mu}_{0}\bigr)^{\!\top}\mSigma_{0}^{-1}\bigl(\bm{\lambda}_{n}-\bm{\mu}_{0}\bigr)\,+\,\bm{\lambda}_{n}^{\top}\bm{G}_{n}\,+\,\mathrm{const.}
    \end{equation}
    Expanding the quadratic and collecting terms in $\bm{\lambda}_{n}$ yields a quadratic with precision matrix $\mSigma_{0}^{-1}$ (unchanged because the likelihood is linear in $\bm{\lambda}_{n}$) and linear term $\mSigma_{0}^{-1}\bm{\mu}_{0}+\bm{G}_{n}$. Completing the square recovers a Gaussian with mean
    \begin{equation}
      \bm{\mu}^{\star}_{n}\,=\,\mSigma_{0}\!\bigl(\mSigma_{0}^{-1}\bm{\mu}_{0}+\bm{G}_{n}\bigr)\,=\,\bm{\mu}_{0}+\mSigma_{0}\bm{G}_{n}
    \end{equation}
    and covariance $\mSigma_{0}$, as claimed.

    \subsection{Sum-Product Belief Propagation on the Simplified Factor Graph}
    \label{appx-subsec: posterior-bp}

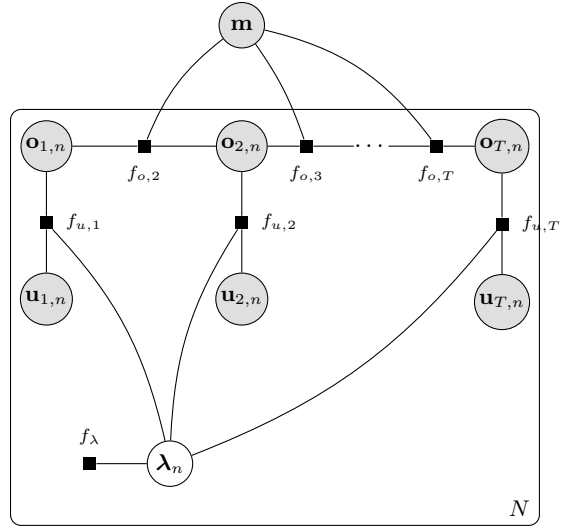
\begin{figure}[!t]
    \centering
    \begin{tikzpicture}[
        scale=0.95, transform shape,
        latent/.style={circle, fill=white, draw=black, inner sep=1pt, minimum size=18pt, font=\fontsize{9}{9}\selectfont, node distance=1},
        obs/.style={latent, fill=gray!25},
        const/.style={rectangle, inner sep=0pt, node distance=1},
        factor/.style={rectangle, fill=black, minimum size=5pt, inner sep=0pt, node distance=0.4},
        flabel/.style={font=\scriptsize, node distance=0},
        edge/.style={-, black},
    ]
        \node[obs]                                                    (o1)    {$\rvo_{1,n}$};
        \node[obs, right=20mm of o1]                                  (o2)    {$\rvo_{2,n}$};
        \node[const, right=12mm of o2]                                (dots)  {$\dots$};
        \node[obs, right=12mm of dots]                                (oT)    {$\rvo_{T,n}$};
        \node[obs, above=10mm of o2]                                  (m)     {$\rvm$};

        \node[factor]                                                 (fo2)   at ($(o1)!0.5!(o2)$) {};
        \node[flabel, below=0.5mm of fo2]                                     {$f_{o,2}$};
        \node[factor]                                                 (fo3)   at ($(o2)!0.5!(dots)$) {};
        \node[flabel, below=0.5mm of fo3]                                     {$f_{o,3}$};
        \node[factor]                                                 (foT)   at ($(dots)!0.5!(oT)$) {};
        \node[flabel, below=0.5mm of foT]                                     {$f_{o,T}$};

        \node[obs, below=14mm of o1]                                  (u1)    {$\rvu_{1,n}$};
        \node[obs, below=14mm of o2]                                  (u2)    {$\rvu_{2,n}$};
        \node[obs, below=14mm of oT]                                  (uT)    {$\rvu_{T,n}$};
        \node[factor]                                                 (fu1)   at ($(o1)!0.5!(u1)$) {};
        \node[flabel, right=0.5mm of fu1]                                     {$f_{u,1}$};
        \node[factor]                                                 (fu2)   at ($(o2)!0.5!(u2)$) {};
        \node[flabel, right=0.5mm of fu2]                                     {$f_{u,2}$};
        \node[factor]                                                 (fuT)   at ($(oT)!0.5!(uT)$) {};
        \node[flabel, right=0.5mm of fuT]                                     {$f_{u,T}$};

        \node[latent, below=16mm of u2, xshift=-1cm]                               (lambda){$\bm{\lambda}_{n}$};
        \node[factor, left=7mm of lambda]                             (flam)  {};
        \node[flabel, above=0.5mm of flam]                                    {$f_{\lambda}$};

        \draw[edge] (o1) -- (fo2); \draw[edge] (fo2) -- (o2);
        \draw[edge] (o2) -- (fo3); \draw[edge] (fo3) -- (dots);
        \draw[edge] (dots) -- (foT); \draw[edge] (foT) -- (oT);
        \draw[edge] (m) to[bend right=15] (fo2);
        \draw[edge] (m) to[bend left=8] (fo3);
        \draw[edge] (m) to[bend left=20] (foT);

        \draw[edge] (o1) -- (fu1); \draw[edge] (fu1) -- (u1);
        \draw[edge] (o2) -- (fu2); \draw[edge] (fu2) -- (u2);
        \draw[edge] (oT) -- (fuT); \draw[edge] (fuT) -- (uT);
        \draw[edge] (lambda) to[bend right=15]  (fu1);
        \draw[edge] (lambda) to[bend left=15]  (fu2);
        \draw[edge] (lambda) to[bend right=15] (fuT);
        \draw[edge] (flam) -- (lambda);

        \node[wrap={(o1)(o2)(oT)(u1)(u2)(uT)(lambda)(fo2)(fo3)(foT)(fu1)(fu2)(fuT)(flam)}] (N-wrap) {};
        \node[plate caption=N-wrap]                                                       (N-caption) {$N$};
        \node[plate={(N-wrap)(N-caption)}]                                                (N)         {};
    \end{tikzpicture}
    \caption{Simplified factor graph corresponding to the relaxed joint~\eqref{eq: simplified-generative}: the latent state $\bm{s}_{1:T}$ and its transition kernel are marginalised away (\cref{sec: problem}), leaving an autoregressive observation chain conditioned on the static map $\bm{m}$. Filled squares are factors, circles are random variables (shaded if observed). The three factor families are: the autoregressive emission factors $f_{o,t}(\bm{o}_{t,n}\!\mid\!\bm{o}_{<t,n},\bm{m})$ (modelled by the $\bm{\lambda}$-conditional CFM), the time-discounted optimality factors $f_{u,t}$ (\cref{appx-subsec: posterior-bp}), and the behavior-latent prior $f_{\lambda}$. Sum-product belief propagation on the per-agent subgraph rooted at $\bm{\lambda}_{n}$ recovers the analytic conjugate posterior of~\Cref{prop: conjugate}.}
    \label{appx-fig: factor-graph}
\end{figure}

    The conjugate posterior is also recovered exactly by sum-product message passing on the tree-structured per-agent factor subgraph induced by the relaxed joint~\eqref{eq: simplified-generative}; the simplified factor graph is shown in~\Cref{appx-fig: factor-graph}. The per-step un-normalized factors $\psi_{t}$ of~\eqref{eq: psi-factor} attach directly to the trajectory--latent pair; the geometric time-discount $\gamma^{t-1}$ is folded into the factor itself, so no separate ``discounted optimality'' assumption is needed beyond the definition~\eqref{eq: psi-factor}.

    \Cref{appx-fig: factor-graph} makes the three factor families that remain after relaxation explicit: the autoregressive emission factors $f_{o,t}(\vo_{t,n}\!\mid\!\vo_{<t,n},\vm)$, the un-normalized optimality factors $\{\psi_{t}\}_{t=1}^{T}$, and the behavior-latent prior $f_{\lambda}(\bm{\lambda}_{n})\!=\!p_{0}(\bm{\lambda}_{n})$. Each filled square is one factor of the unnormalized joint~\eqref{eq: simplified-generative}, and each circle is a random variable (shaded if observed). Because the latent state $\vs_{1:T}$ and its transitions are absent, the per-agent optimality chains decouple immediately and sum-product is exact on each agent's subgraph.

    \noindent\textbf{Message from a single optimality factor.} Because the observation $\vo_{t,n}$ is observed, the incoming message from the $\vo_{t,n}$ node to $\psi_{t}$ is the Dirac indicator $\delta(\vo_{t,n}\!-\!\vo_{t,n}^{\mathrm{obs}})$. The outbound message that $\psi_{t}$ sends to $\bm{\lambda}_{n}$ is the factor evaluated at the logged observation,
    \begin{equation}
      m_{\psi_{t}\to\bm{\lambda}_{n}}\bigl(\bm{\lambda}_{n}\bigr)\,=\,\psi_{t}(\vo_{t,n}^{\mathrm{obs}},\bm{\lambda}_{n})\,=\,\exp\!\bigl(\gamma^{t-1}\,\bm{\lambda}_{n}^{\top}\bm{r}(\vo_{t,n}^{\mathrm{obs}})\bigr).
      \label{eq: psi-message}
    \end{equation}
    The factor $\psi_{t}$ is un-normalized on the $\bm{\lambda}_{n}$--$\vo_{t,n}$ pair, so no additional summation step is introduced at the factor node.

    \noindent\textbf{Product across time.} The $T$ outbound messages meet at the $\bm{\lambda}_{n}$ variable node, where sum-product multiplies them:
    \begin{equation}
      \prod_{t=1}^{T}\,m_{\psi_{t}\to\bm{\lambda}_{n}}\bigl(\bm{\lambda}_{n}\bigr)\,=\,\exp\!\Bigl(\,\bm{\lambda}_{n}^{\top}\sum_{t=1}^{T}\gamma^{t-1}\,\bm{r}(\vo_{t,n}^{\mathrm{obs}})\Bigr)\,=\,\exp\!\bigl(\bm{\lambda}_{n}^{\top}\bm{G}_{n}\bigr),
      \label{eq: bp-product}
    \end{equation}
    which reproduces~\eqref{eq: cumret} of the main text. The geometric discount enters the BP product exclusively through the per-step factor~\eqref{eq: psi-factor}; if instead $\gamma\!=\!1$ were chosen (time-homogeneous factor), the same algebra yields the undiscounted sum $\bm{\lambda}_{n}^{\top}\sum_{t}\bm{r}(\vo_{t,n})$.

    \noindent\textbf{Combination with the prior factor.} The variable node $\bm{\lambda}_{n}$ also receives a single inbound message from the behavior-prior factor $f_{\lambda}(\bm{\lambda}_{n})\!=\!p_{0}(\bm{\lambda}_{n})\!=\!\gN(\bm{\mu}_{0},\mSigma_{0})$. The belief at $\bm{\lambda}_{n}$ is the product of all inbound messages:
    \begin{equation}
      b(\bm{\lambda}_{n})\,\propto\,f_{\lambda}(\bm{\lambda}_{n})\,\prod_{t=1}^{T}\,m_{\psi_{t}\to\bm{\lambda}_{n}}\bigl(\bm{\lambda}_{n}\bigr)\,\propto\,\exp\!\Bigl(\,-\tfrac{1}{2}(\bm{\lambda}_{n}-\bm{\mu}_{0})^{\top}\mSigma_{0}^{-1}(\bm{\lambda}_{n}-\bm{\mu}_{0})\,+\,\bm{\lambda}_{n}^{\top}\bm{G}_{n}\Bigr).
    \end{equation}
    Completing the square (as in~\cref{appx-subsec: posterior-completing-square}) gives
    \begin{equation}
      b(\bm{\lambda}_{n})\,=\,\gN\!\bigl(\bm{\mu}_{0}+\mSigma_{0}\bm{G}_{n},\,\mSigma_{0}\bigr)\,=\,q^{\star}(\bm{\lambda}_{n}),
    \end{equation}
    which is the analytic posterior of~\Cref{prop: conjugate}. Because the subgraph rooted at $\bm{\lambda}_{n}$, comprising $f_{\lambda}$ and the $T$ factors $\{\psi_{t}\}_{t=1}^{T}$, is a tree, sum-product is exact and the belief $b(\bm{\lambda}_{n})$ coincides with the true marginal posterior. The remaining factors of the joint (initial-state prior, transition, observation) are independent of $\bm{\lambda}_{n}$ given the observed trajectory, so their contributions appear only in the normalization constant.

    Because the latent-state marginalization in~\eqref{eq: simplified-generative} has already removed $\vs_{1:T}$ from the joint, the per-agent optimality chains in~\Cref{appx-fig: factor-graph} are independent across $n$, and sum-product is exact agent-by-agent.

    \section{Extended Method Details}
    \label{appx-sec: extended-method}

    This appendix collects the return-labeling and eligibility details deferred from~\cref{sec: method}: the context-residual and lane-centerline return definitions, the soft-eligibility decay formulas, and the contrastive conditioning augmentation.

    \subsection{Context-Residual Return Labeling}
    \label{appx-subsec: context-residual-detail}

    The raw per-channel return $G_{n,k}$ conflates agent behavior with scenario difficulty: a highway scene yields low offroad penalties regardless of driving style. We define the \emph{context-residual return} as the per-channel discounted return residualized against the scene context,
    \begin{equation}
      G^{\mathrm{cr}}_{n,k}\;\triangleq\;G_{n,k}\;-\;\bar{G}_{k}(\vm_{n}),
      \label{eq: context-residual}
    \end{equation}
    where $\bar{G}_{k}(\vm_{n})$ is the mean return for agents sharing the same map context $\vm_{n}$, estimated once on the calibration split. After standardization via~\eqref{eq: g-standardise}, $G^{\mathrm{cr}}_{n,k}$ isolates the behavioral signal from the structural component, producing posterior labels that are more informative for conditioning. The context-residual formulation is the primary return measure used for all CNeVA training and evaluation in~\cref{sec: experiment}.

    \subsection{Lane-Centerline Map Reward}
    \label{appx-subsec: lane-centerline-detail}

    As an alternative to the composite offroad-plus-road-edge penalty used in the context-residual return, we define a \emph{lane-centerline return} based on the signed lateral offset to the nearest lane centerline:
    \begin{equation}
      r^{\mathrm{lc}}_{t,n}\;\triangleq\;-\,\bigl|d^{\perp}_{t,n}\bigr|\,/\,w^{\mathrm{lane}}_{t,n},
      \label{eq: lane-centerline-reward}
    \end{equation}
    where $d^{\perp}_{t,n}$ is the perpendicular distance from agent $n$ at time $t$ to the nearest lane centerline and $w^{\mathrm{lane}}_{t,n}$ is the corresponding lane half-width. This measure provides a purely geometric, coordinate-specific definition of map compliance. Below, we evaluate the map CSM diagonal under this measure alongside the context-residual and physical-offroad alternatives.

    \subsection{Soft Eligibility Gates: Decay Formulas}
    \label{appx-subsec: soft-eligibility-detail}

    Under the hard-gated baseline, safety labels are computed only for agents whose minimum clearance $c_{n}\!<\!5\,\text{m}$ or minimum time-to-collision $\mathrm{ttc}_{n}\!<\!6\,\text{s}$; map labels only for agents within a road-boundary margin or off-road. This binary gating introduces a discontinuity at the threshold and excludes the majority of agents from safety/map supervision. We replace the hard gates with smooth exponential decay applied to the per-step pairwise risk. For safety, each pairwise collision risk term is modulated by both clearance and TTC:
    \begin{equation}
      \tilde{r}^{\mathrm{safe}}_{t,ij}\;\triangleq\;r^{\mathrm{safe}}_{t,ij}\cdot\exp\!\Bigl(-\frac{\max(c_{t,ij},\,0)}{\tau_{c}}\Bigr)\cdot\exp\!\Bigl(-\frac{\max(\mathrm{ttc}_{t,ij},\,0)}{\tau_{t}}\Bigr),
      \label{eq: soft-safety}
    \end{equation}
    with decay scales $\tau_{c}\!=\!2.0\,\text{m}$ and $\tau_{t}\!=\!3.0\,\text{s}$. The product form ensures that both clearance \emph{and} TTC must be large for the risk to vanish: a distant but fast-approaching agent (low TTC, high clearance) retains a high risk weight. For map compliance, the per-step risk is modulated by the signed margin $m_{t,n}$ to the nearest road boundary:
    \begin{equation}
      \tilde{r}^{\mathrm{map}}_{t,n}\;\triangleq\;r^{\mathrm{map}}_{t,n}\cdot\exp\!\Bigl(-\frac{\max(m_{t,n},\,0)}{\tau_{m}}\Bigr),
      \label{eq: soft-map}
    \end{equation}
    with $\tau_{m}\!=\!1.0\,\text{m}$. Under soft eligibility, all valid agents receive labels (the eligibility condition reduces to validity alone), but agents far from hazards contribute negligibly to the discounted return because the decay drives the per-step reward toward zero. Still, agents

    \subsection{Contrastive Conditioning}
    \label{appx-subsec: contrastive-detail}

    The mixed channel-mask curriculum of~\cref{subsec: cfm} exposes the unconditional and conditional paths during training, but does not explicitly contrast them within the same batch. We augment training with a \emph{contrastive conditioning} step: for each agent $n$, the loss includes both the steered forward pass (with $\bm{\lambda}_{n}\!\sim\!q^{\star}$) and the null forward pass (with $\ve_{\varnothing}$), computed over the same flow-time noise realization. This paired evaluation encourages the velocity field to learn the \emph{difference} between steered and unsteered trajectories rather than their absolute positions, thereby improving the robustness of the conditioning signal during closed-loop rollout, where distribution shift would otherwise attenuate the guidance effect. The drift-paired CSM evaluation protocol of~\cref{subsec: exp-controllability} mirrors this training-time pairing at test time.

    \section{Model Architecture and Training Details}
    \label{appx-sec: model-details}

    This appendix collects the parametric and operational details deferred from~\cref{subsec: vi,subsec: cfm,subsec: inference}: the calibration of the prior covariance $\mSigma_{0}$, the $\bm{\lambda}$-conditional rectified-flow decoder architecture, the classifier-free mixed-mask curriculum, and the deployment-time sampler. Training uses~\eqref{eq: cfm-cfg-loss} on a v4-32 slice with batch $4$ per device, AdamW~\citep{loshchilov2019decoupled} at peak learning rate $10^{-4}$, $2000$ warm-up steps, and weight decay $10^{-2}$.

    \subsection{Per-Channel Return Standardization}
    \label{appx-subsec: sigma-calibration}

\begin{figure}[h]
      \centering
      \includegraphics[width=\linewidth]{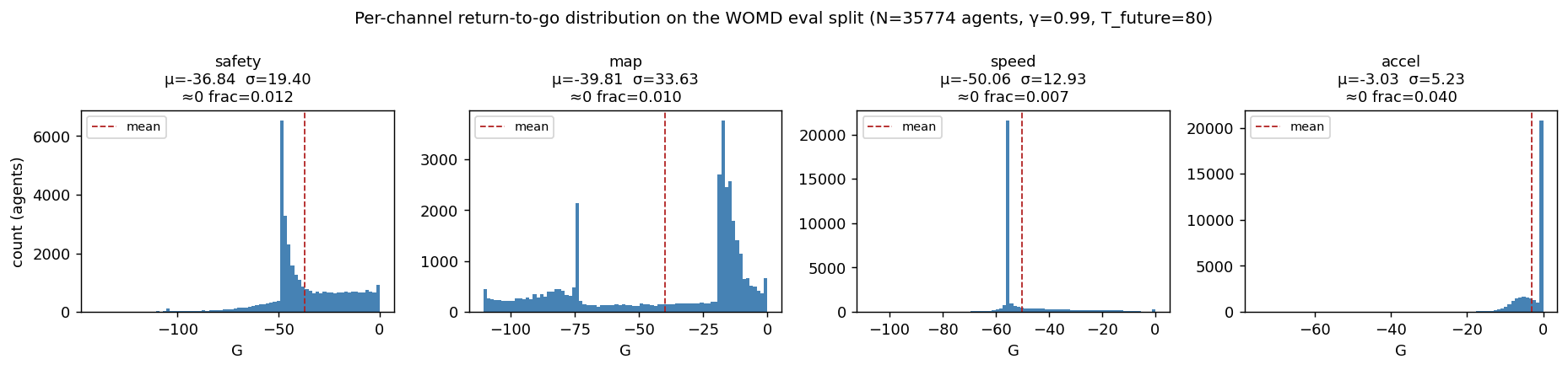}
      \caption{Empirical histogram of the per-channel discounted return $G_{n,k}$ over the fixed WOMD calibration split (N=35,774 simulated agents, $\gamma\!=\!0.99$, $T_{f}\!=\!80$). Per-channel mean and standard deviation are annotated; the dashed red line marks the mean. The four channels differ in scale by up to an order of magnitude (speed mean $\approx\!-50$, accel mean $\approx\!-3$), motivating the per-channel standardization of~\eqref{eq: g-stats}. The map channel is visibly bimodal: most agents stay on-map, while a tail of extended off-road episodes drags the mean down.}
      \label{appx-fig: rtg-hist}
    \end{figure}

    The raw per-channel discounted return $\bm{G}_{n}$ on WOMD demonstrations has substantially different scales across the four channels: the safety channel mean is around $-37$, map around $-40$, speed around $-50$, and accel only around $-3$; per-channel standard deviations span $5.2$ (accel) to $33.6$ (map). Feeding $\bm{G}_{n}$ verbatim into the conjugate update of~\eqref{eq: analytic-posterior} with $\mSigma_{0}\!=\!\bm{I}$ would let the speed and map channels dominate the per-agent $\bm{\lambda}_{n}$ by an order of magnitude, making the acceleration channel difficult for the conditional generator to use. We therefore standardize per channel using the empirical statistics measured on a fixed WOMD calibration split with $\gamma\!=\!0.99$ and $T_{f}\!=\!80$:
    \begin{equation}
      \bm{\mu}_{G}\,\approx\,\bigl[-36.84,\,-39.81,\,-50.06,\,-3.03\bigr],\qquad
      \bm{\sigma}_{G}\,\approx\,\bigl[19.40,\,33.63,\,12.93,\,5.23\bigr],
      \label{eq: g-stats}
    \end{equation}
    in channel order $(\text{safety},\text{map},\text{speed},\text{accel})$. The standardized return $\widetilde{G}_{n,k}\!=\!(G_{n,k}-\mu_{G,k})/\sigma_{G,k}$ replaces $G_{n,k}$ inside~\eqref{eq: analytic-posterior} and~\eqref{eq: reparam}. With this preconditioning the prior covariance reduces to the only free hyperparameter; we use $\bm{\mu}_{0}\!=\!\bm{0}$ and $\mSigma_{0}\!=\!\bm{I}$. The empirical histograms underlying~\eqref{eq: g-stats} are shown in~\Cref{appx-fig: rtg-hist}; the bimodal shape of the map channel (with a heavy left tail from extended off-road episodes) motivated keeping the constituent offroad and dist-to-road-edge terms together rather than separating them into two reward channels.

    \subsection{Reward-Channel Construction}
    \label{appx-subsec: reward-channels}
    The reward basis used for posterior labeling is implemented as four sign-aligned penalty sums. The safety channel is collision plus distance-to-nearest-object plus time-to-collision shaping; the map channel is offroad plus distance-to-road-edge; the speed channel is linear plus angular speed; and the accel channel is linear plus angular acceleration. Each constituent is shaped into $[-1,0]$ before discounting, and a backward scan accumulates the per-channel discounted return $G_{n,k}$. Agents without a valid future horizon are masked out of the CFM loss, return-standardization statistics, and steering metrics; their reward tensors are zero-filled only after this masking step, so invalid agents do not contribute best-possible zero reward labels. Positive steering of channel $k$ therefore means reducing the corresponding penalty mass.
    \subsection{Lambda-Conditional Rectified-Flow Decoder}
    \label{appx-subsec: cfm-architecture}

    The velocity field $\bm{v}_{\vtheta}(\vx_{s},s,\vo_{<t,n},\vm,\bm{\lambda}_{n})$ is a Transformer decoder that consumes (i) the noisy displacement latent $\vx_{s}$, (ii) the flow time $s$ embedded by Fourier features (sampled as sigmoid of a standard normal in implementation), (iii) the cross-attention conditioning set, and produces the predicted velocity $\bm{v}^{\star}_{n}$. The conditioning set
    \begin{equation}
      \mathrm{cond}_{n}\,\triangleq\,\bigl[\,\mathrm{Dense}_{K\!\to\!d_{h}}(\bm{\lambda}_{n});\;\mathrm{embed}(\vo_{<t,n});\;\mathrm{embed}(\vm)\,\bigr]
      \label{eq: cond-set}
    \end{equation}
    folds the per-agent latent into an extra token alongside the road-graph and traffic-light embeddings, in the same way per-agent map context is consumed by BehaviorGPT~\citep{zhou2024behaviorgpt}. Each decoder block stacks (a) a temporally causal self-attention over the future axis, (b) an unmasked spatial self-attention across agents, (c) a cross-attention to $\mathrm{cond}_{n}$, and (d) a feed-forward sub-block, each preceded by an adaptive layer-norm whose scale and shift parameters are predicted from the flow-time embedding.
    For masked conditioning, the implementation concatenates the value-zeroed lambda vector with the binary channel mask before projection, giving a 2K-dimensional token input; Eq.~(20) shows the unmasked special case.

    \subsection{Mixed Channel-Mask CFG Curriculum}
    \label{appx-subsec: classifier-free}

    \noindent\textbf{Why not classifier guidance?} A naive alternative would be to train an unconditional velocity $\bm{v}_{\vtheta}^{\mathrm{un}}(\vx_{s},s,\,\cdot\,)$ together with a separate regressor $p_{\bm{\phi}}(\bm{G}\!\mid\!\vx_{s},s)$ on noised trajectories and steer sampling by $\nabla_{\vx_{s}}\!\log p_{\bm{\phi}}$, transposing the \emph{classifier guidance} recipe of~\citet{dhariwal2021adm} from class labels to continuous returns. Two practical issues carry over. First, $p_{\bm{\phi}}$ must be trained on the same diffused-trajectory distribution as the decoder, and its supervisory signal at large $s$, where $\vx_{s}$ is nearly isotropic Gaussian noise, is essentially uninformative. Meanwhile, stepping along $\nabla_{\vx_{s}}\!\log p_{\bm{\phi}}$ in a continuous, high-dimensional $\bm{\lambda}$-space resembles a gradient-based adversarial attack on $p_{\bm{\phi}}$~\citep{ho2022cfg}: high-confidence directions of the regressor are not the same as directions of high data density.

    \noindent\textbf{Why a curriculum?} The vanilla classifier-free guidance recipe~\citep{ho2022cfg} flips between fully-conditional ($\bm{\lambda}\!=\!\bm{\lambda}_{n}$) and fully-unconditional ($\bm{\lambda}\!=\!\ve_{\varnothing}$) updates with a single Bernoulli mask. At inference, the operator-supplied one-hot probe $\bm{\lambda}\!=\!\rho\bm{e}_{k}$, with only channel $k$ informative and the remaining $K\!-\!1$ at value zero, is then strictly outside the training support. As a result, the network has only ever seen dense $\bm{\lambda}$ drawn from $q^{\star}$. We therefore expose every \emph{subset} of channels during training by drawing $u_{n}\!\sim\!\gU(0,1)$ and setting
    \begin{equation}
      \vb_{n} \,\sim\,
      \begin{cases}
        \bm{1}_{K} & 0\,\le\,u_{n}\,<\,0.2 \quad (\text{null path; learned null token wins})\\
        \mathrm{Mask}_{1}(K) & 0.2\,\le\,u_{n}\,<\,0.6 \quad (\text{exactly one channel kept})\\
        \mathrm{Mask}_{2}(K) & 0.6\,\le\,u_{n}\,<\,0.8 \quad (\text{exactly two channels kept})\\
        \bm{0}_{K} & 0.8\,\le\,u_{n}\,\le\,1.0\quad (\text{all channels kept}),
      \end{cases}
      \label{eq: mask-curriculum}
    \end{equation}
    where $\vb_{n,k}\!=\!1$ marks channel $k$ as masked-out and $\vb_{n,k}\!=\!0$ marks it as kept, and $\mathrm{Mask}_{j}(K)$ is the uniform distribution over binary masks with exactly $j$ kept entries. The one-hot probe at channel $k$ now matches the single-channel training distribution exactly (mask everywhere except $k$).

    \noindent\textbf{Mask indicator concat.} For the projection to distinguish ``$\lambda_{k}\!=\!0$ (observed)'' from ``$\lambda_{k}$ unobserved,'' we concatenate the value-zeroed profile with the mask indicator before the $\R^{2K}\!\to\!\R^{d_{h}}$ Dense projection:
    \begin{equation}
      \widetilde{\bm{\lambda}}_{n}\,\triangleq\,\bigl[(\bm{1}-\vb_{n})\odot\bm{\lambda}_{n};\;\vb_{n}\bigr]\,\in\,\R^{2K}.
      \label{eq: lambda-tilde-supp}
    \end{equation}
    When $\vb_{n}\!=\!\bm{1}_{K}$ (the $20\%$ null branch) we additionally override the projection output with the learned null embedding $\ve_{\varnothing}\!\in\!\R^{d_{h}}$, recovering the unconditional path of Ho--Salimans. The conditional and unconditional velocity heads share all backbone parameters except $\ve_{\varnothing}$. The training loop is summarized in~\Cref{appx-alg: training}.

    \begin{algorithm}[H]
    \caption{Training}
    \label{appx-alg: training}
    \begin{algorithmic}[1]
    \REQUIRE Dataset $\gD$.
    \REQUIRE Reward channel count $K$; mask probabilities $(p_{0},p_{1},p_{2},p_{K})\!=\!(0.20,0.40,0.20,0.20)$.
    \REQUIRE Standardisation statistics $\bm{\mu}_{G},\bm{\sigma}_{G}$; prior $(\bm{\mu}_{0},\mSigma_{0})$.

    \REPEAT
      \STATE Sample a scenario $(\vo_{1:T,1:N},\vm)\!\sim\!\gD$.
      \FORALL{agent $n\!=\!1,\dots,N$}
        \STATE Compute $\widetilde{G}_{n,k}\!=\!(G_{n,k}-\mu_{G,k})/\sigma_{G,k}$ for each $k$.\hfill$\triangleright$ standardised return
        \STATE Sample $\bm{\eta}_{n}\!\sim\!\gN(\bm{0},\bm{I})$ and set $\bm{\lambda}_{n}\!=\!\bm{\mu}_{0}+\mSigma_{0}\widetilde{\bm{G}}_{n}+\mSigma_{0}^{1/2}\bm{\eta}_{n}$.
        \STATE Sample $u_{n}\!\sim\!\gU(0,1)$ and choose $j\!\in\!\{0,1,2,K\}$ with probabilities $(p_{0},p_{1},p_{2},p_{K})$.
        \STATE Sample a uniformly random size-$j$ subset $\gK_{n}\!\subseteq\!\{1,\dots,K\}$
        \FORALL{$k\!=\!1,\dots,K$}
            \STATE Set $\vb_{n,k}\!=\!\mathbb{I}\{k\!\not\in\!\gK_{n}\}$.\hfill$\triangleright$ channel mask
        \ENDFOR
        \STATE Form $\widetilde{\bm{\lambda}}_{n}\!=\![(\bm{1}-\vb_{n})\odot\bm{\lambda}_{n};\;\vb_{n}]\!\in\!\R^{2K}$; if $j\!=\!0$ flag agent $n$ as null.
        \STATE Sample $s$ as sigmoid of a standard normal and sample epsilon from a standard normal.
        \STATE Form $\vx_{s}\!=\!(1-s)\bm{y}_{n}+s\bm{\epsilon}$ and the rectified-flow target $\bm{v}^{\star}_{n}\!=\!\bm{\epsilon}-\bm{y}_{n}$.
      \ENDFOR
      \STATE Inside $\bm{v}_{\vtheta}$, replace the Dense projection of $\widetilde{\bm{\lambda}}_{n}$ by $\ve_{\varnothing}$ for every agent flagged null.
      \STATE $\vtheta\!\leftarrow\!\vtheta-\alpha\,\nabla_{\vtheta}\!\sum_{n}\bigl\Vert\bm{v}_{\vtheta}(\vx_{s},s,\vo_{<t,n},\vm,\widetilde{\bm{\lambda}}_{n})-\bm{v}^{\star}_{n}\bigr\Vert^{2}$.
    \UNTIL{converged}
    \end{algorithmic}
    \end{algorithm}

    \noindent\textbf{Score-form correspondence.} The CFG combination~\eqref{eq: cfg-velocity} is the rectified-flow analogue of the noise-prediction form $\widetilde{\bm{\epsilon}}\!=\!(1\!+\!w)\bm{\epsilon}_{\vtheta}(\bm{z},\bm{c})\!-\!w\bm{\epsilon}_{\vtheta}(\bm{z},\ve_{\varnothing})$ used by~\citet{ho2022cfg}: the rectified-flow velocity is an affine reparameterization of the score $\nabla_{\vx_{s}}\!\log p_{\vtheta}(\vx_{s}\!\mid\!\bm{\lambda})$ under the linear noise schedule of~\eqref{eq: rectified-flow}~\citep{lipman2023flow}, so the linear extrapolation in velocity space inherits the implicit-classifier interpretation $\nabla_{\vx_{s}}\!\log p^{i}(\bm{\lambda}\!\mid\!\vx_{s})\!=\!\nabla_{\vx_{s}}\!\log p(\vx_{s}\!\mid\!\bm{\lambda})\!-\!\nabla_{\vx_{s}}\!\log p(\vx_{s})$ of~\citet{ho2022cfg}; the resulting sample is approximately drawn from $p(\vx_{s}\!\mid\!\bm{\lambda})\,p^{i}(\bm{\lambda}\!\mid\!\vx_{s})^{w}$.

    \subsection{Rectified-Flow Euler Sampler with Classifier-Free Guidance}
    \label{appx-subsec: euler-sampler}

    Deployment recovers a trajectory from $\vx_{1}\!\sim\!\gN(\bm{0},\bm{I})$ by Euler-integrating the rectified-flow ODE in reverse time, with the bare conditional velocity replaced by the CFG combination of~\eqref{eq: cfg-velocity}:
    \begin{equation}
      \vx_{s-\Delta s}\,\leftarrow\,\vx_{s}\,-\,\Delta s\;\widetilde{\bm{v}}_{\vtheta}^{\,w}(\vx_{s},\bm{\lambda}_{n}^{\mathrm{op}}),\qquad s\!\in\!\{1,1\!-\!\Delta s,\dots,\Delta s\},
      \label{eq: euler}
    \end{equation}
    where $\widetilde{\bm{v}}_{\vtheta}^{\,w}$ is evaluated by two forward passes of $\bm{v}_{\vtheta}$ with conditioning $\bm{\lambda}_{n}^{\mathrm{op}}$ and $\ve_{\varnothing}$, respectively. Activating the learned null token for every agent recovers the unconditional population marginal; any other choice of $(\bm{\lambda}_{n}^{\mathrm{op}},w)$ steers the trajectory along the requested preference direction without modifying the integrator or the noise schedule. The receding-horizon variant of~\cref{subsec: inference} simply re-evaluates~\eqref{eq: euler} every $\ell$ output timesteps with a refreshed conditioning history. \Cref{appx-alg: sampling} states the full sampler.

    \begin{algorithm}[H]
    \caption{One-Shot Sampling}
    \label{appx-alg: sampling}
    \begin{algorithmic}[1]
    \REQUIRE Operator preference $\bm{\lambda}_{n}^{\mathrm{op}}$.
    \REQUIRE Guidance scale $w\!\geq\!0$; integrator steps $T_{\mathrm{int}}$.
    \REQUIRE History $\vo_{<t,n}$ and map $\vm$.
    \STATE Sample $\vx_{1}\!\sim\!\gN(\bm{0},\bm{I})$; let $\Delta s\!=\!1/T_{\mathrm{int}}$.
    \FOR{$s\!=\!1,1\!-\!\Delta s,\dots,\Delta s$}
      \STATE $\bm{v}^{\mathrm{c}}\!\leftarrow\!\bm{v}_{\vtheta}(\vx_{s},s,\vo_{<t,n},\vm,\bm{\lambda}_{n}^{\mathrm{op}})$\hfill$\triangleright$ conditional pass
      \STATE $\bm{v}^{\mathrm{u}}\!\leftarrow\!\bm{v}_{\vtheta}(\vx_{s},s,\vo_{<t,n},\vm,\ve_{\varnothing})$\hfill$\triangleright$ unconditional pass
      \STATE $\widetilde{\bm{v}}\!\leftarrow\!(1\!+\!w)\,\bm{v}^{\mathrm{c}}\,-\,w\,\bm{v}^{\mathrm{u}}$\hfill$\triangleright$ CFG combination
      \STATE $\vx_{s-\Delta s}\!\leftarrow\!\vx_{s}-\Delta s\,\widetilde{\bm{v}}$
    \ENDFOR
    \RETURN $\vx_{0}$
    \end{algorithmic}
\end{algorithm}

    The receding-horizon deployment regime of~\cref{subsec: inference} wraps~\cref{appx-alg: sampling} in a fixed-stride patch loop: every $\ell$ output timesteps, the model re-conditions on a history buffer that has been updated with its own previous predictions, so the CFG double forward is repeated within each patch and the conditioning is refreshed at every patch boundary.

    \begin{algorithm}[H]
    \caption{Receding-Horizon Patch Sampling}
    \label{appx-alg: patch-sampler}
    \begin{algorithmic}[1]
    \REQUIRE Operator preference $\bm{\lambda}_{n}^{\mathrm{op}}$.
    \REQUIRE Guidance scale $w\!\geq\!0$ and integrator steps $T_{\mathrm{int}}$
    \REQUIRE History $\vo_{<1,n}$; map $\vm$; horizon $T$; patch length $\ell$.
    \STATE Initialise the history buffer $\vo^{\mathrm{hist}}\!\leftarrow\!\vo_{<1,n}$.
    \FOR{$t_{\mathrm{start}}\!=\!1, 1\!+\!\ell, 1\!+\!2\ell,\dots,T\!-\!\ell\!+\!1$}
      \STATE Run~\cref{appx-alg: sampling} with conditioning $\vo^{\mathrm{hist}}$ to obtain the next patch of $\ell$ displacement predictions $\hat{\bm{y}}_{t_{\mathrm{start}}:t_{\mathrm{start}}+\ell-1}$.
      \STATE Integrate displacements into per-step observations $\hat{\vo}_{t_{\mathrm{start}}:t_{\mathrm{start}}+\ell-1}$ via $\hat{\vo}_{t}\!=\!\hat{\vo}_{t-1}\!+\!\hat{\bm{y}}_{t}$.
      \STATE Append $\hat{\vo}_{t_{\mathrm{start}}:t_{\mathrm{start}}+\ell-1}$ to $\vo^{\mathrm{hist}}$.
    \ENDFOR
    \RETURN concatenated $\hat{\vo}_{1:T}$
    \end{algorithmic}
\end{algorithm}

    Setting $\ell\!=\!T$ recovers the single-pass open-loop sampler; setting $\ell\!=\!1$ corresponds to per-step replan. We use $\ell\!=\!16$ throughout, matching the training-time patch length used by the CFM loss in~\eqref{eq: cfm-cfg-loss}.

    \section{Pairwise Trajectory Divergence}
    \label{appx-sec: divergence}
    \suppressfloats[t]
    \Cref{appx-tab: divergence} expands the pairwise trajectory-divergence summary from the main text. It confirms that speed and acceleration induce the largest geometric separation, while the safety-map pair remains near the noise floor, matching the weak safety/map CSM response.
\renewcommand{\divergencemean}{0.54}
\begin{table}[t]
  \centering
  \small
  \setlength{\tabcolsep}{6pt}
  \caption{Pairwise trajectory divergence $D(\bm{e}_{a},\bm{e}_{b})$ at $\rho\!=\!1$, $w\!=\!1.5$ (m). Paired flow-time noise; single-channel-kept mask. Mean over pairs $\bar{D}\!=\!0.54$\,m.}
  \label{appx-tab: divergence}
  \begin{tabular}{lcccc}
    \toprule
     & \textbf{safety} & \textbf{map} & \textbf{speed} & \textbf{accel} \\
    \midrule
    \textbf{safety} & -- & 0.10 & 0.63 & 0.49 \\
    \textbf{map} &  & -- & 0.62 & 0.47 \\
    \textbf{speed} &  &  & -- & 0.96 \\
    \textbf{accel} &  &  &  & -- \\
    \bottomrule
  \end{tabular}
\end{table}

    \section{Supplementary Diagnostics}
    \label{appx-sec: supplementary-diagnostics}
    This section provides additional multi-seed CSM diagnostics for the model in~\cref{sec: experiment}, thereby expanding the aggregated results in~\cref{tab: csm-main}. These diagnostics illustrate the controllability hierarchy discussed in~\cref{sec: experiment}.

    \subsection{Interpretation and Limitations}
    \label{appx-subsec: diagnostic-interpretation}

    These diagnostics support the limitation stated in the main text: CNeVA's controllability depends strongly on the chosen reward basis and the identifiability of behavior labels under reconstruction training. Dense kinematic channels such as speed and acceleration yield consistent geometric responses. In contrast, safety and map compliance are sparse, context-dependent, and harder to capture with one-hot $\bm{\lambda}$ steering. The main model addresses the safety aspect of this limitation through soft eligibility gates (\cref{subsec: soft-eligibility}), which yield a statistically significant safety CSM, whereas the hard-eligibility ablation does not (\cref{tab: plausibility}). Map controllability remains a structural limitation of the current reward basis.

    \subsection{Conditioning Ablation}
    \label{appx-subsec: conditioning-ablation}

    As a minimal ablation on the model, we compare the steered CNeVA path against the unconditional path ($\bm{\lambda}\!=\!\ve_{\varnothing}$) under identical flow-time noise; the CSM diagonals reported in~\cref{tab: csm-main} are the result. Removing the conditioning entirely (i.e., zeroing $\bm{\lambda}$ and the mask indicator) recovers the unconditional baseline, confirming that the steering effect is attributable to the behavior latent and not to the noise realization.

    \begin{figure}[t]
      \centering
      \includegraphics[width=\linewidth]{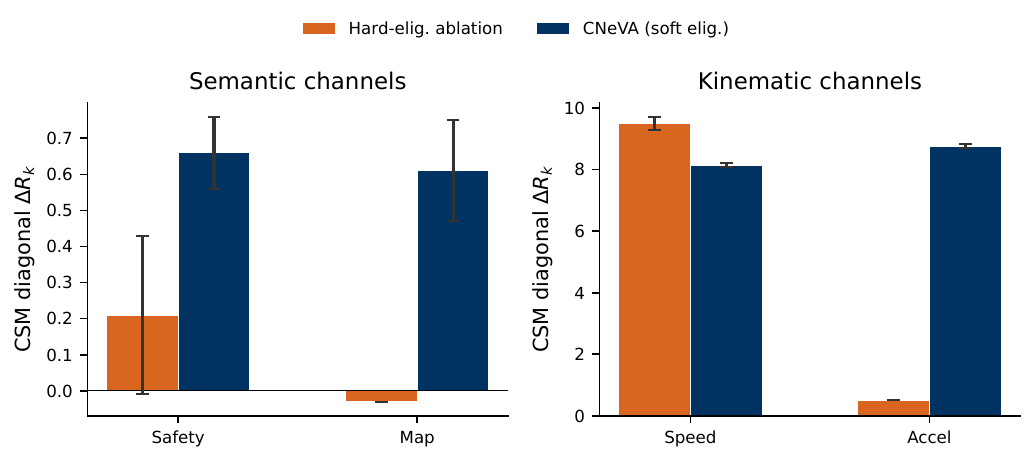}
      \caption{CSM diagonal $\Delta R_k$ for CNeVA versus the hard-eligibility ablation at $\rho\!=\!1$, $w\!=\!1.5$ (open-loop, context-residual return). Left: semantic channels (safety, map). Right: kinematic channels (speed, accel). Error bars show $\pm 1$ std over $5$ seeds.}
      \label{fig: controllability-overview}
    \end{figure}

    \subsection{Multi-Seed CSM}
    \label{appx-subsec: multi-seed-csm}

    \Cref{tab: csm-main} in the main text reports the CSM diagonals aggregated over $5$ evaluation seeds ($s\!=\!42$--$46$). The per-seed standard deviations are small relative to the between-channel differences for the kinematic channels (speed std $0.07$; accel std $0.07$) and moderate for the semantic channels (safety std $0.10$; map std $0.14$). The unconditional baseline is near-zero across all seeds (safety: $+0.06 \pm 0.18$; map: $+0.06 \pm 0.06$), confirming that the null path has no directional steering bias.

    \subsection{Physical Plausibility Diagnostics}
    \label{appx-subsec: plausibility-table}

    \Cref{tab: plausibility} reports the physical-plausibility diagnostics for the main model and the two hard-eligibility ablations discussed in~\cref{subsec: exp-plausibility}: the stall fraction (per-step displacement $<0.1$\,m), the mean speed of steered agents as a fraction of logged ground-truth speed, and the offroad rate at $8$\,s under speed steering, alongside the speed and safety CSM diagonals. The early-stage ablation's inflated speed CSM (${+}51.3$) is achieved by stalling ($75.9\%$ of agents, $61\%$ of GT speed), whereas the main model retains $94.7\%$ of GT speed at a physically valid ${+}8.15$; the hard-eligibility ablation's safety CSM collapses to ${+}0.21$ versus the main model's ${+}0.66$.

    \begin{table}[!t]
      \centering
      \small
      \setlength{\tabcolsep}{3.5pt}
      \caption{Physical plausibility: CNeVA with soft eligibility versus two hard-eligibility ablations. Open-loop, $\rho\!=\!1$, $w\!=\!1.5$, mean $\pm$ std over $5$ seeds. Soft and hard eligibility define different $\bm{\lambda}$ populations, so safety CSM values are not directly commensurable across rows.}
      \label{tab: plausibility}
      \begin{tabular}{lccccccc}
        \toprule
        & & \multicolumn{2}{c}{\textbf{Speed steering}} & & & \\
        \cmidrule(lr){3-4}
        \textbf{Model} & \textbf{minADE}$\downarrow$ & Stall\%$\downarrow$ & $v/v_{\mathrm{GT}}$$\uparrow$ & $\Delta R_{\mathrm{speed}}$ & $\Delta R_{\mathrm{safety}}$ & Offroad\%$\downarrow$ \\
        \midrule
        \textbf{CNeVA} & $1.113 \pm .011$ & $65.1$ & $94.7\%$ & ${+}8.15 \pm 0.07$ & ${+}0.66 \pm 0.10$ & $32.5$ \\
        \midrule
        \multicolumn{7}{l}{\emph{Ablations (hard eligibility)}} \\
        \quad Early-stage (40K) & $1.238 \pm .009$ & $75.9$ & $61\%$  & ${+}51.3 \pm 0.3$ & ${+}2.18 \pm 0.10$ & $34.6$ \\
        \quad Hard-elig.\ (200K) & $1.112 \pm .015$ & $65.4$ & $94\%$ & ${+}9.5 \pm 0.2$  & ${+}0.21 \pm 0.22$ & $32.9$ \\
        \bottomrule
      \end{tabular}
    \end{table}

    \subsection{Map Controllability Across Reward Measures}
    \label{appx-subsec: map-reward-ablation}

\begin{figure}[h]
      \centering
      \includegraphics[width=0.6\linewidth]{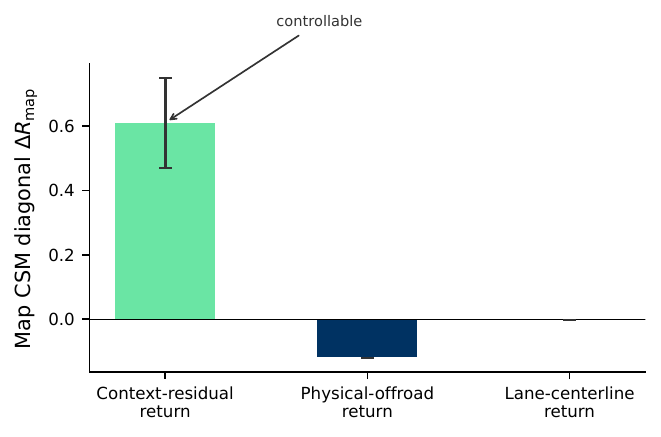}
      \caption{Map CSM diagonal ($\Delta R_{\mathrm{map}}$) across three return measures, evaluated on the 200K soft-eligibility model at $\rho\!=\!1$, $w\!=\!1.5$ (open-loop). Map steering is controllable only under the context-residual definition ($+0.61$); the physical-offroad ($-0.12$) and lane-centerline (${\approx}\,0$) measures show no measurable physical-space response.}
      \label{appx-fig: map-three-space}
    \end{figure}

    This section evaluates the map CSM diagonal under two return definitions alternative to the context-residual measure: the \emph{physical-offroad return} (raw non-residualized offroad plus road-edge penalty) and the \emph{lane-centerline return} (\cref{appx-subsec: lane-centerline-detail}). \Cref{appx-fig: map-three-space} shows a stark sensitivity. Under the context-residual return the map diagonal is positive ($\Delta R_{\mathrm{map}}\!=\!{+}0.61 \pm 0.14$); under the physical-offroad return it is slightly negative ($-0.12$); and under the lane-centerline return it is effectively zero ($-0.002$). The context-residual measure residualizes away the scenario-dependent baseline, so the remaining variation is behaviorally attributable; the coordinate-specific measures do not, and the model's map-channel response fails to separate from the scenario effect. The result has two implications. First, map controllability in~\cref{tab: csm-main} is genuine but measure-dependent, reflecting a shift relative to the scene-contextual expectation rather than absolute geometric compliance. Meanwhile, coordinate-level map control (\emph{e.g., lane-keeping in a specific lane}) would require richer reward decompositions that separate spatial from temporal map compliance.

    \putbib[references]

\end{appendices}

\end{bibunit}


\begin{thebibliography}{39}
\providecommand{\natexlab}[1]{#1}
\providecommand{\url}[1]{\texttt{#1}}
\expandafter\ifx\csname urlstyle\endcsname\relax
  \providecommand{\doi}[1]{doi: #1}\else
  \providecommand{\doi}{doi: \begingroup \urlstyle{rm}\Url}\fi

\bibitem[Cao et~al.(2024)Cao, Ivanovic, Xiao, and Pavone]{cao2024rlhf}
Yulong Cao, Boris Ivanovic, Chaowei Xiao, and Marco Pavone.
\newblock Reinforcement learning with human feedback for realistic traffic
  simulation.
\newblock In \emph{2024 IEEE International Conference on Robotics and
  Automation (ICRA)}, pages 14428--14434, May 2024.
\newblock \doi{10.1109/ICRA57147.2024.10610878}.

\bibitem[Chang et~al.(2023)Chang, Tang, Li, Hu, Tomizuka, and
  Zhan]{chang2023socialedit}
Wei-Jer Chang, Chen Tang, Chenran Li, Yeping Hu, Masayoshi Tomizuka, and Wei
  Zhan.
\newblock Editing driver character: Socially-controllable behavior generation
  for interactive traffic simulation.
\newblock \emph{IEEE Robotics and Automation Letters}, 8\penalty0 (9):\penalty0
  5432--5439, September 2023.
\newblock \doi{10.1109/LRA.2023.3291897}.

\bibitem[Chang et~al.(2024)Chang, Pittaluga, Tomizuka, Zhan, and
  Chandraker]{chang2024safesim}
Wei-Jer Chang, Francesco Pittaluga, Masayoshi Tomizuka, Wei Zhan, and Manmohan
  Chandraker.
\newblock {SAFE-SIM}: Safety-critical closed-loop traffic simulation with
  diffusion-controllable adversaries.
\newblock In \emph{European Conference on Computer Vision (ECCV)}, 2024.

\bibitem[Chang et~al.(2026)Chang, Rangesh, Joseph, Strong, Tomizuka, Hu, and
  Zhan]{chang2026spacer}
Wei-Jer Chang, Akshay Rangesh, Kevin Joseph, Matthew Strong, Masayoshi
  Tomizuka, Yihan Hu, and Wei Zhan.
\newblock {SPACeR}: Self-play anchoring with centralized reference models.
\newblock In \emph{International Conference on Learning Representations
  (ICLR)}, 2026.

\bibitem[Cornelisse and Vinitsky(2024)]{cornelisse2024hrppo}
Daphne Cornelisse and Eugene Vinitsky.
\newblock Human-compatible driving partners through data-regularized self-play
  reinforcement learning, 2024.

\bibitem[Cusumano-Towner et~al.(2025)Cusumano-Towner, Hafner, Hertzberg, Huval,
  Petrenko, Vinitsky, Wijmans, Killian, Bowers, Sener, Kraehenbuehl, and
  Koltun]{cusumano2025gigaflow}
Marco Cusumano-Towner, David Hafner, Alexander Hertzberg, Brody Huval, Aleksei
  Petrenko, Eugene Vinitsky, Erik Wijmans, Taylor Killian, Stuart Bowers, Ozan
  Sener, Philipp Kraehenbuehl, and Vladlen Koltun.
\newblock Robust autonomy emerges from self-play.
\newblock In \emph{Proceedings of the 42nd International Conference on Machine
  Learning}, volume 267 of \emph{PMLR}, pages 11710--11737, 2025.

\bibitem[Ettinger et~al.(2021)Ettinger, Cheng, Caine, Liu, Zhao, Pradhan, Chai,
  Sapp, Qi, Zhou, et~al.]{ettinger2021womd}
Scott Ettinger, Shuyang Cheng, Benjamin Caine, Chenxi Liu, Hang Zhao, Sabeek
  Pradhan, Yuning Chai, Benjamin Sapp, Charles~R Qi, Yin Zhou, et~al.
\newblock Large scale interactive motion forecasting for autonomous driving:
  The {W}aymo open motion dataset.
\newblock In \emph{IEEE/CVF International Conference on Computer Vision
  (ICCV)}, 2021.

\bibitem[Ho and Salimans(2022)]{ho2022cfg}
Jonathan Ho and Tim Salimans.
\newblock Classifier-free diffusion guidance, 2022.

\bibitem[Hu et~al.(2025)Hu, Chai, Yang, Qian, Li, Shao, Zhang, Xu, and
  Liu]{hu2024gump}
Yihan Hu, Siqi Chai, Zhening Yang, Jingyu Qian, Kun Li, Wenxin Shao, Haichao
  Zhang, Wei Xu, and Qiang Liu.
\newblock Solving motion planning tasks with a scalable generative model.
\newblock In Ale{\v{s}} Leonardis, Elisa Ricci, Stefan Roth, Olga Russakovsky,
  Torsten Sattler, and G{\"u}l Varol, editors, \emph{Computer Vision -- ECCV
  2024}, pages 386--404, Cham, 2025. Springer Nature Switzerland.

\bibitem[Huang et~al.(2026)Huang, Zhang, Vaidya, Chen, Fernández~Fisac, and
  Lv]{huang2024vbd}
Zhiyu Huang, Zixu Zhang, Ameya Vaidya, Yuxiao Chen, Jaime Fernández~Fisac, and
  Chen Lv.
\newblock Versatile behavior diffusion for generalized traffic agent
  simulation.
\newblock \emph{IEEE Transactions on Intelligent Transportation Systems}, pages
  1--17, 2026.

\bibitem[Jiang et~al.(2023)Jiang, Cornman, Park, Sapp, Zhou, and
  Anguelov]{jiang2023motiondiffuser}
Chiyu~Max Jiang, Andre Cornman, Cheolho Park, Benjamin Sapp, Yin Zhou, and
  Dragomir Anguelov.
\newblock {MotionDiffuser}: Controllable multi-agent motion prediction using
  diffusion.
\newblock In \emph{IEEE/CVF Conference on Computer Vision and Pattern
  Recognition (CVPR)}, 2023.

\bibitem[Jiang et~al.(2024)Jiang, Bai, Cornman, Davis, Huang, Jeon,
  Kulshrestha, Lambert, Li, Zhou, Fuertes, Yuan, Tan, Zhou, and
  Anguelov]{jiang2024scenediffuser}
Chiyu~Max Jiang, Yijing Bai, Andre Cornman, Christopher Davis, Xiukun Huang,
  Hong Jeon, Sakshum Kulshrestha, John Lambert, Shuangyu Li, Xuanyu Zhou,
  Carlos Fuertes, Chang Yuan, Mingxing Tan, Yin Zhou, and Dragomir Anguelov.
\newblock Scenediffuser: Efficient and controllable driving simulation
  initialization and rollout.
\newblock In \emph{Advances in Neural Information Processing Systems},
  volume~37, pages 55729--55760, 2024.

\bibitem[Jordan(1999)]{jordan1999learning}
Michael~Irwin Jordan.
\newblock \emph{Learning in graphical models}.
\newblock MIT press, 1999.

\bibitem[Lipman et~al.(2023)Lipman, Chen, Ben-Hamu, Nickel, and
  Le]{lipman2023flow}
Yaron Lipman, Ricky T~Q Chen, Heli Ben-Hamu, Maximilian Nickel, and Matt Le.
\newblock Flow matching for generative modeling.
\newblock In \emph{International Conference on Learning Representations
  (ICLR)}, 2023.

\bibitem[Liu et~al.(2026)Liu, Liu, Zhang, and Huang]{liu2026stage}
Zihao Liu, Xing Liu, Yizhai Zhang, and Panfeng Huang.
\newblock Stage: Style-controllable action generation for personalized
  autonomous driving.
\newblock \emph{IEEE Robotics and Automation Letters}, 11\penalty0
  (2):\penalty0 2130--2137, February 2026.
\newblock \doi{10.1109/LRA.2025.3640974}.

\bibitem[Montali et~al.(2023)Montali, Lambert, Mougin, Kuefler, Rhinehart, Li,
  Gulino, Emrich, Yang, Whiteson, et~al.]{montali2023wosac}
Nico Montali, John Lambert, Paul Mougin, Alex Kuefler, Nicholas Rhinehart,
  Michelle Li, Cole Gulino, Tristan Emrich, Zoey Yang, Shimon Whiteson, et~al.
\newblock The {W}aymo open sim agents challenge.
\newblock In \emph{Advances in Neural Information Processing Systems (NeurIPS)
  Datasets and Benchmarks Track}, 2023.

\bibitem[Nayakanti et~al.(2023)Nayakanti, Al-Rfou, Zhou, Goel, Refaat, and
  Sapp]{nayakanti2023wayformer}
Nigamaa Nayakanti, Rami Al-Rfou, Aurick Zhou, Kratarth Goel, Khaled~S Refaat,
  and Benjamin Sapp.
\newblock Wayformer: Motion forecasting via simple \& efficient attention
  networks.
\newblock In \emph{IEEE International Conference on Robotics and Automation
  (ICRA)}, 2023.

\bibitem[Ngiam et~al.(2022)Ngiam, Caine, Vasudevan, Zhang, Chiang, Ling,
  Roelofs, Bewley, Liu, Venugopal, Weiss, Sapp, Chen, and
  Shlens]{ngiam2022scenetransformer}
Jiquan Ngiam, Benjamin Caine, Vijay Vasudevan, Zhengdong Zhang, Hao-Tien~Lewis
  Chiang, Jeffrey Ling, Rebecca Roelofs, Alex Bewley, Chenxi Liu, Ashish
  Venugopal, David Weiss, Ben Sapp, Zhifeng Chen, and Jonathon Shlens.
\newblock Scene transformer: A unified architecture for predicting multiple
  agent trajectories, 2022.

\bibitem[Pei et~al.(2026)Pei, Shi, and
  Shen]{pei2026advancingmultiagenttrafficsimulation}
Muleilan Pei, Shaoshuai Shi, and Shaojie Shen.
\newblock Advancing multi-agent traffic simulation via r1-style reinforcement
  fine-tuning, 2026.

\bibitem[Philion et~al.(2024)Philion, Peng, and Fidler]{philion2024trajeglish}
Jonah Philion, Xue~Bin Peng, and Sanja Fidler.
\newblock {Trajeglish}: Traffic modeling as next-token prediction.
\newblock In \emph{International Conference on Learning Representations
  (ICLR)}, 2024.

\bibitem[Pronovost et~al.(2023)Pronovost, Ganesina, Hendy, Wang, Morales, Wang,
  and Roy]{Pronovost2023scenariodiffusion}
Ethan Pronovost, Meghana~Reddy Ganesina, Noureldin Hendy, Zeyu Wang, Andres
  Morales, Kai Wang, and Nick Roy.
\newblock Scenario diffusion: Controllable driving scenario generation with
  diffusion.
\newblock In \emph{Advances in Neural Information Processing Systems
  (NeurIPS)}, 2023.

\bibitem[Rempe et~al.(2022)Rempe, Philion, Guibas, Fidler, and
  Litany]{rempe2022accidentprone}
Davis Rempe, Jonah Philion, Leonidas~J. Guibas, Sanja Fidler, and Or~Litany.
\newblock Generating useful accident-prone driving scenarios via a learned
  traffic prior.
\newblock In \emph{IEEE/CVF Conference on Computer Vision and Pattern
  Recognition (CVPR)}, 2022.

\bibitem[Rowe et~al.(2024)Rowe, Girgis, Gosselin, Carrez, Golemo, Heide, Paull,
  and Pal]{rowe2024ctrlsim}
Luke Rowe, Roger Girgis, Anthony Gosselin, Bruno Carrez, Florian Golemo, Felix
  Heide, Liam Paull, and Christopher Pal.
\newblock {CtRL-Sim}: Reactive and controllable driving agents with offline
  reinforcement learning.
\newblock In \emph{Conference on Robot Learning (CoRL)}, 2024.

\bibitem[Rowe et~al.(2025)Rowe, Girgis, Gosselin, Paull, Pal, and
  Heide]{rowe2025scenariodreamer}
Luke Rowe, Roger Girgis, Anthony Gosselin, Liam Paull, Christopher Pal, and
  Felix Heide.
\newblock Scenario dreamer: Vectorized latent diffusion for generating driving
  simulation environments.
\newblock In \emph{IEEE/CVF Conference on Computer Vision and Pattern
  Recognition (CVPR)}, 2025.

\bibitem[Seff et~al.(2023)Seff, Cera, Chen, Ng, Zhou, Nayakanti, Refaat,
  Al-Rfou, and Sapp]{seff2023motionlm}
Ari Seff, Brian Cera, Dian Chen, Mason Ng, Aurick Zhou, Nigamaa Nayakanti,
  Khaled~S Refaat, Rami Al-Rfou, and Benjamin Sapp.
\newblock {MotionLM}: Multi-agent motion forecasting as language modeling.
\newblock In \emph{IEEE/CVF International Conference on Computer Vision
  (ICCV)}, 2023.

\bibitem[Shi et~al.(2024)Shi, Jiang, Dai, and Schiele]{shi2024mtrpp}
Shaoshuai Shi, Li~Jiang, Dengxin Dai, and Bernt Schiele.
\newblock Mtr++: Multi-agent motion prediction with symmetric scene modeling
  and guided intention querying.
\newblock \emph{IEEE Transactions on Pattern Analysis and Machine
  Intelligence}, 46\penalty0 (5):\penalty0 3955--3971, May 2024.
\newblock \doi{10.1109/TPAMI.2024.3352811}.

\bibitem[Suo et~al.(2021)Suo, Regalado, Casas, and Urtasun]{suo2021trafficsim}
Simon Suo, Sebastian Regalado, Sergio Casas, and Raquel Urtasun.
\newblock {TrafficSim}: Learning to simulate realistic multi-agent behaviors.
\newblock In \emph{IEEE/CVF Conference on Computer Vision and Pattern
  Recognition (CVPR)}, 2021.

\bibitem[Tan et~al.(2025{\natexlab{a}})Tan, Lambert, Jeon, Kulshrestha, Bai,
  Luo, Anguelov, Tan, and Jiang]{tan2025scenediffuser2}
Shuhan Tan, John Lambert, Hong Jeon, Sakshum Kulshrestha, Yijing Bai, Jing Luo,
  Dragomir Anguelov, Mingxing Tan, and Chiyu~Max Jiang.
\newblock {SceneDiffuser++}: City-scale traffic simulation via a generative
  world model.
\newblock In \emph{IEEE/CVF Conference on Computer Vision and Pattern
  Recognition (CVPR)}, 2025{\natexlab{a}}.

\bibitem[Tan et~al.(2025{\natexlab{b}})Tan, Zheng, Liang, Wang, ZHENG, Zheng,
  Li, Zhan, and Liu]{tan2025flowplanner}
Tianyi Tan, Yinan Zheng, Ruiming Liang, Zexu Wang, Kexin ZHENG, Jinliang Zheng,
  Jianxiong Li, Xianyuan Zhan, and Jingjing Liu.
\newblock Flow matching-based autonomous driving planning with advanced
  interactive behavior modeling.
\newblock In \emph{Advances in Neural Information Processing Systems
  (NeurIPS)}, 2025{\natexlab{b}}.

\bibitem[Wu et~al.(2024)Wu, Feng, Gao, and Kan]{wu2024smart}
Wei Wu, Xiaoxin Feng, Ziyan Gao, and Yuheng Kan.
\newblock Smart: Scalable multi-agent real-time motion generation via
  next-token prediction.
\newblock In \emph{Advances in Neural Information Processing Systems},
  volume~37, pages 114048--114071, 2024.

\bibitem[Xing et~al.(2025)Xing, Zhang, Hu, Jiang, He, Zhang, Long, and
  Yin]{xing2025goalflow}
Zebin Xing, Xingyu Zhang, Yang Hu, Bo~Jiang, Tong He, Qian Zhang, Xiaoxiao
  Long, and Wei Yin.
\newblock {GoalFlow}: Goal-driven flow matching for multimodal trajectories
  generation in end-to-end autonomous driving.
\newblock In \emph{IEEE/CVF Conference on Computer Vision and Pattern
  Recognition (CVPR)}, 2025.

\bibitem[Yin et~al.(2021)Yin, Sun, Sun, Tomizuka, and Zhan]{yin2021routegan}
Zhao-Heng Yin, Lingfeng Sun, Liting Sun, Masayoshi Tomizuka, and Wei Zhan.
\newblock Diverse critical interaction generation for planning and planner
  evaluation.
\newblock In \emph{2021 IEEE/RSJ International Conference on Intelligent Robots
  and Systems (IROS)}, pages 7036--7043, September 2021.
\newblock \doi{10.1109/IROS51168.2021.9636266}.

\bibitem[Zhang et~al.(2022)Zhang, Gao, Zhang, Guo, Ding, Wang, Sun, and
  Zhao]{zhang2022trajgen}
Qichao Zhang, Yinfeng Gao, Yikang Zhang, Youtian Guo, Dawei Ding, Yunpeng Wang,
  Peng Sun, and Dongbin Zhao.
\newblock Trajgen: Generating realistic and diverse trajectories with reactive
  and feasible agent behaviors for autonomous driving.
\newblock \emph{IEEE Transactions on Intelligent Transportation Systems},
  23\penalty0 (12):\penalty0 24474--24487, December 2022.
\newblock ISSN 1558-0016.
\newblock \doi{10.1109/TITS.2022.3202185}.

\bibitem[Zhang et~al.(2023)Zhang, Liniger, Dai, Yu, and
  Van~Gool]{zhang2023trafficbots}
Zhejun Zhang, Alexander Liniger, Dengxin Dai, Fisher Yu, and Luc Van~Gool.
\newblock {TrafficBots}: Towards world models for autonomous driving simulation
  and motion prediction.
\newblock In \emph{IEEE International Conference on Robotics and Automation
  (ICRA)}, 2023.

\bibitem[Zhang et~al.(2025{\natexlab{a}})Zhang, Karkus, Igl, Ding, Chen,
  Ivanovic, and Pavone]{zhang2025catk}
Zhejun Zhang, Peter Karkus, Maximilian Igl, Wenhao Ding, Yuxiao Chen, Boris
  Ivanovic, and Marco Pavone.
\newblock Closed-loop supervised fine-tuning of tokenized traffic models.
\newblock In \emph{IEEE/CVF Conference on Computer Vision and Pattern
  Recognition (CVPR)}, 2025{\natexlab{a}}.

\bibitem[Zhang et~al.(2025{\natexlab{b}})Zhang, Jia, Chen, Li, and
  Yan]{trajtok2025}
Zhiyuan Zhang, Xiaosong Jia, Guanyu Chen, Qifeng Li, and Junchi Yan.
\newblock {TrajTok}: Technical report for 2025 {Waymo} open sim agents
  challenge.
\newblock Technical report, Shanghai Jiao Tong University, 2025{\natexlab{b}}.

\bibitem[Zhong et~al.(2023{\natexlab{a}})Zhong, Rempe, Chen, Ivanovic, Cao, Xu,
  Pavone, and Ray]{zhong2023ctgpp}
Ziyuan Zhong, Davis Rempe, Yuxiao Chen, Boris Ivanovic, Yulong Cao, Danfei Xu,
  Marco Pavone, and Baishakhi Ray.
\newblock Language-guided traffic simulation via scene-level diffusion.
\newblock In \emph{Conference on Robot Learning (CoRL)}, 2023{\natexlab{a}}.

\bibitem[Zhong et~al.(2023{\natexlab{b}})Zhong, Rempe, Xu, Chen, Veer, Che,
  Ray, and Pavone]{zhong2023ctg}
Ziyuan Zhong, Davis Rempe, Danfei Xu, Yuxiao Chen, Sushant Veer, Tong Che,
  Baishakhi Ray, and Marco Pavone.
\newblock Guided conditional diffusion for controllable traffic simulation.
\newblock In \emph{2023 IEEE International Conference on Robotics and
  Automation (ICRA)}, pages 3560--3566, 2023{\natexlab{b}}.

\bibitem[Zhou et~al.(2024)Zhou, Hu, Chen, Wang, Guan, Wu, Li, Huang, and
  Xue]{zhou2024behaviorgpt}
Zikang Zhou, Haibo Hu, Xinhong Chen, Jianping Wang, Nan Guan, Kui Wu, Yung-Hui
  Li, Yu-Kai Huang, and Chun~Jason Xue.
\newblock {BehaviorGPT}: Smart agent simulation for autonomous driving with
  next-patch prediction.
\newblock In \emph{Advances in Neural Information Processing Systems
  (NeurIPS)}, 2024.

\end{thebibliography}


\begin{thebibliography}{5}
\providecommand{\natexlab}[1]{#1}
\providecommand{\url}[1]{\texttt{#1}}
\expandafter\ifx\csname urlstyle\endcsname\relax
  \providecommand{\doi}[1]{doi: #1}\else
  \providecommand{\doi}{doi: \begingroup \urlstyle{rm}\Url}\fi

\bibitem[Dhariwal and Nichol(2021)]{dhariwal2021adm}
Prafulla Dhariwal and Alex Nichol.
\newblock Diffusion models beat {GAN}s on image synthesis.
\newblock In \emph{Advances in Neural Information Processing Systems
  (NeurIPS)}, 2021.

\bibitem[Ho and Salimans(2022)]{ho2022cfg}
Jonathan Ho and Tim Salimans.
\newblock Classifier-free diffusion guidance, 2022.

\bibitem[Lipman et~al.(2023)Lipman, Chen, Ben-Hamu, Nickel, and
  Le]{lipman2023flow}
Yaron Lipman, Ricky T~Q Chen, Heli Ben-Hamu, Maximilian Nickel, and Matt Le.
\newblock Flow matching for generative modeling.
\newblock In \emph{International Conference on Learning Representations
  (ICLR)}, 2023.

\bibitem[Loshchilov and Hutter(2019)]{loshchilov2019decoupled}
Ilya Loshchilov and Frank Hutter.
\newblock Decoupled weight decay regularization.
\newblock In \emph{International Conference on Learning Representations
  (ICLR)}, 2019.
\newblock URL \url{https://openreview.net/forum?id=Bkg6RiCqY7}.

\bibitem[Zhou et~al.(2024)Zhou, Hu, Chen, Wang, Guan, Wu, Li, Huang, and
  Xue]{zhou2024behaviorgpt}
Zikang Zhou, Haibo Hu, Xinhong Chen, Jianping Wang, Nan Guan, Kui Wu, Yung-Hui
  Li, Yu-Kai Huang, and Chun~Jason Xue.
\newblock {BehaviorGPT}: Smart agent simulation for autonomous driving with
  next-patch prediction.
\newblock In \emph{Advances in Neural Information Processing Systems
  (NeurIPS)}, 2024.

\end{thebibliography}
\end{document}